\documentclass[numbers]{article}

    \usepackage[final]{neurips_2025}

\usepackage[utf8]{inputenc} 
\usepackage[T1]{fontenc}    
\usepackage{hyperref}       
\hypersetup{colorlinks, breaklinks, citecolor=[rgb]{0, 0.44, 0.737},
anchorcolor=[rgb]{1, 0.44, 0.737}, linkcolor=[rgb]{0, 0.44, 0.737},
urlcolor=[rgb]{0, 0.44, 0.737} 
}
\usepackage{url}            
\usepackage{booktabs}       
\usepackage{amsfonts}       
\usepackage{nicefrac}       
\usepackage{microtype}      
\usepackage[dvipsnames]{xcolor}         
\usepackage{amsmath}
\usepackage{graphicx}
\usepackage{colortbl}
\usepackage{amssymb}
\usepackage{multirow}
\usepackage[font=small,labelfont=bf,tableposition=top]{caption}
\DeclareCaptionLabelFormat{andtable}{#1~#2  \&  \tablename~\thetable}
\usepackage{pifont}
\usepackage{subfig}
\usepackage{rotating}
\usepackage{tablefootnote}
\usepackage{enumitem}
\usepackage{makecell}
\usepackage{wrapfig}
\usepackage{float}
\newcommand{\cmark}{\ding{52}}%
\newcommand{\xmark}{\ding{56}}%
\newcommand{\ie}{\textit{i.e.}}
\newcommand{\eg}{\textit{e.g.}}
\newcommand{\etc}{\textit{etc}}

\title{OpenWorldSAM: Extending SAM2 for Universal Image Segmentation with Language Prompts}

\author{%
  Shiting Xiao\\
  Yale University\\
  {\small \texttt{ginny.xiao@yale.edu}}
  \And
  Rishabh Kabra\\
  Google DeepMind\\
  {\small \texttt{rkabra@google.com}}
  \And
  Yuhang Li\\
  Yale University\\
  {\small \texttt{yuhang.li@yale.edu}}
\And
Donghyun Lee\\
Yale University \\
 {\small \texttt{donghyun.lee@yale.edu}}\\
\And
Jo\~ao Carreira\\
Google DeepMind\\
{\small \texttt{joaoluis@google.com}}
\And
Priyadarshini Panda \\
Yale University \\
{\small \texttt{priya.panda@yale.edu}}
}

\begin{document}

\maketitle

\vspace{-15pt}
\begin{abstract}
The ability to segment objects based on open-ended language prompts remains a critical challenge, requiring models to ground textual semantics into precise spatial masks while handling diverse and unseen categories. We present OpenWorldSAM, a framework that extends the prompt-driven Segment Anything Model v2 (SAM2) to open-vocabulary scenarios by integrating multi-modal embeddings extracted from a lightweight vision-language model (VLM).
  Our approach is guided by four key principles: i) \textit{Unified prompting}: OpenWorldSAM supports a diverse range of prompts, including category-level and sentence-level language descriptions, providing a flexible interface for various segmentation tasks. ii) \textit{Efficiency}: By freezing the pre-trained components of SAM2 and the VLM, we train only 4.5 million parameters on the COCO-stuff dataset, achieving remarkable resource efficiency.
  iii) \textit{Instance Awareness}: We enhance the model's spatial understanding through positional tie-breaker embeddings and cross-attention layers, enabling effective segmentation of multiple instances. 
  iv) \textit{Generalization}: OpenWorldSAM exhibits strong zero-shot capabilities, generalizing well on unseen categories and an open vocabulary of concepts without additional training. Extensive experiments demonstrate that OpenWorldSAM achieves state-of-the-art performance in open-vocabulary semantic, instance, and panoptic segmentation across multiple benchmarks. Code is available at \href{https://github.com/GinnyXiao/OpenWorldSAM}{https://github.com/GinnyXiao/OpenWorldSAM}.
\end{abstract}

\vspace{-15pt}
\section{Introduction}

Image segmentation has long been constrained to closed-vocabulary settings, where models can only recognize objects from a predefined taxonomy~\cite{long2015fcn, ronneberger2015unet, chen2017deeplab, chen2018deeplab, he2017maskrcnn, xie2021segformer,  cheng2022mask2former, li2023mask}. However, real-world applications, \eg,  Embodied AI~\cite{fan2024active, laina2025findanything},  demand systems that can understand open-ended language descriptions (from single nouns like “pedestrian” to rich referring expressions such as “the man in a red shirt”) and segment novel objects unseen during training. This \textit{open-vocabulary segmentation} problem poses two core challenges: (1) \textit{Semantic grounding} – mapping free-form text to visual entities, and (2) \textit{Instance awareness} – distinguishing multiple objects that match the same description.

Detection-centric methods~\cite{ghiasi2022openseg,liang2023ovseg} relied on two-stage pipelines, first detecting class-agnostic mask proposals then classifying them with vision-language models (VLMs), \eg, CLIP~\cite{radford2021clip} and ALIGN~\cite{jia2021align}. While effective, such approaches struggle with complex queries and specialize exclusively in semantic segmentation, lacking versatility. Recent generalist models~\cite{zou2023xdecoder,shen2024ape}  explore unified architectures that jointly handle vision and language, allowing a single model to perform detection, segmentation, and grounding tasks. These generalist models demonstrate impressive flexibility, but they typically involve resource-intensive pre-training. The emergence of  promptable segmentation models like the Segment Anything Model (SAM)~\cite{kirillov2023sam,ravi2024sam2}  offered new possibilities – it introduced a paradigm shift by allowing users to segment arbitrary objects using simple visual prompts (\eg, points, boxes). Trained on an extensive dataset, these models exhibit remarkable generalization and interactive capabilities. However, they inherently lack semantic understanding. Subsequent attempts to combine SAM with large language models (LLMs)~\cite{lai2024lisa, rasheed2024glamm, xu2024ullava} achieved language awareness but at prohibitive computational costs, imposing overwhelming overhead.

\begin{figure}
    \centering
    \includegraphics[width=\textwidth]{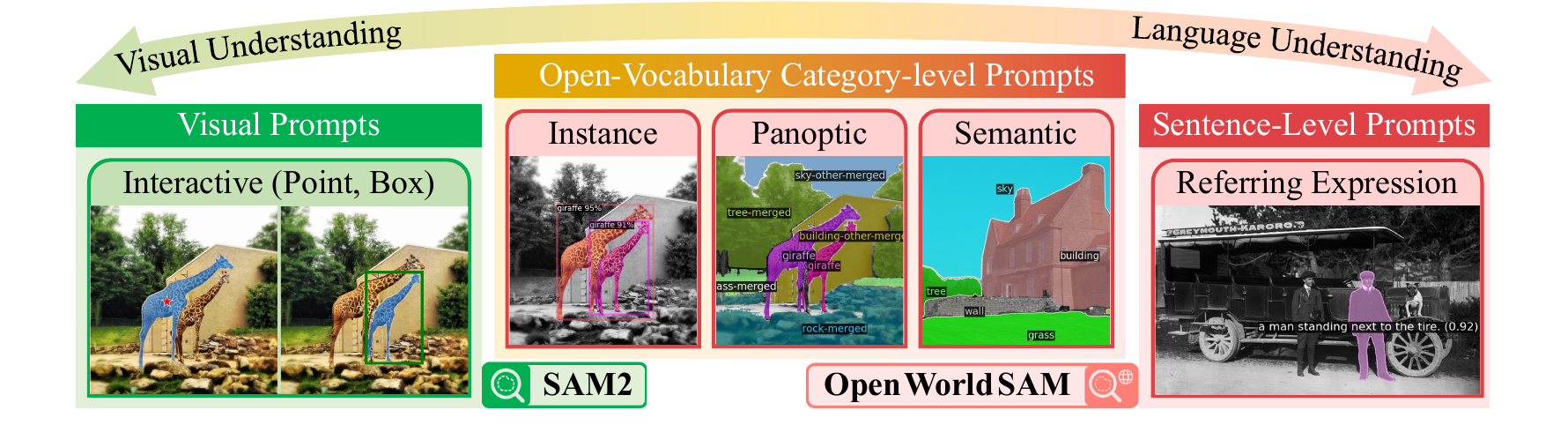}
    \caption{Overview of the proposed framework. The green region highlights the SAM v2 baseline, supporting visual prompts (e.g., boxes, points) for interactive segmentation. Our OpenWorldSAM extension integrates open-vocabulary language understanding, enabling both category-level segmentation across semantic, instance, panoptic tasks and referring expression segmentation.}
    \label{fig:intro}
    \vspace{-17pt}
\end{figure}

We posit that an ideal open-vocabulary segmenter should: (i) Natively support textual prompts without cascaded classification components, (ii) Preserve the knowledge of the vision foundation models like SAM without adding large overhead, and (iii) Segment multiple possible instances that could correspond to a single query. To this end, we propose OpenWorldSAM, an open-vocabulary extension to the SAM v2 (SAM2) architecture that satisfies these requirements. OpenWorldSAM injects language understanding while retaining SAM2’s core strengths through a lightweight language adapter ($\approx$4.5M trainable parameters), unifing category-level instance, semantic, and panoptic segmentation, and sentence-level referring expression segmentation (Figure~\ref{fig:intro}). 

Specifically, we feed the image and descriptive text input into a frozen multi-modal encoder and obtain fused semantic representations. These serve as prompts to SAM2' mask decoder that produces masks for any described object or region. We introduce a positional tie-breaker mechanism to resolve ambiguities when a text query could apply to multiple regions, allowing the model to perform multi-instance segmentation. Furthermore, our adapter employs a soft prompting technique that uses cross-attention between textual queries and image features, sharpening localization by allowing semantic contexts to focus toward relevant image areas.
 By combining these design innovations, OpenWorldSAM can accurately identify and segment arbitrary objects described by text, all while using only frozen pre-trained encoders and a tiny trainable adaptation module.

\begin{wrapfigure}{r}{0.45\textwidth}
  \begin{center}
    \includegraphics[width=\linewidth]{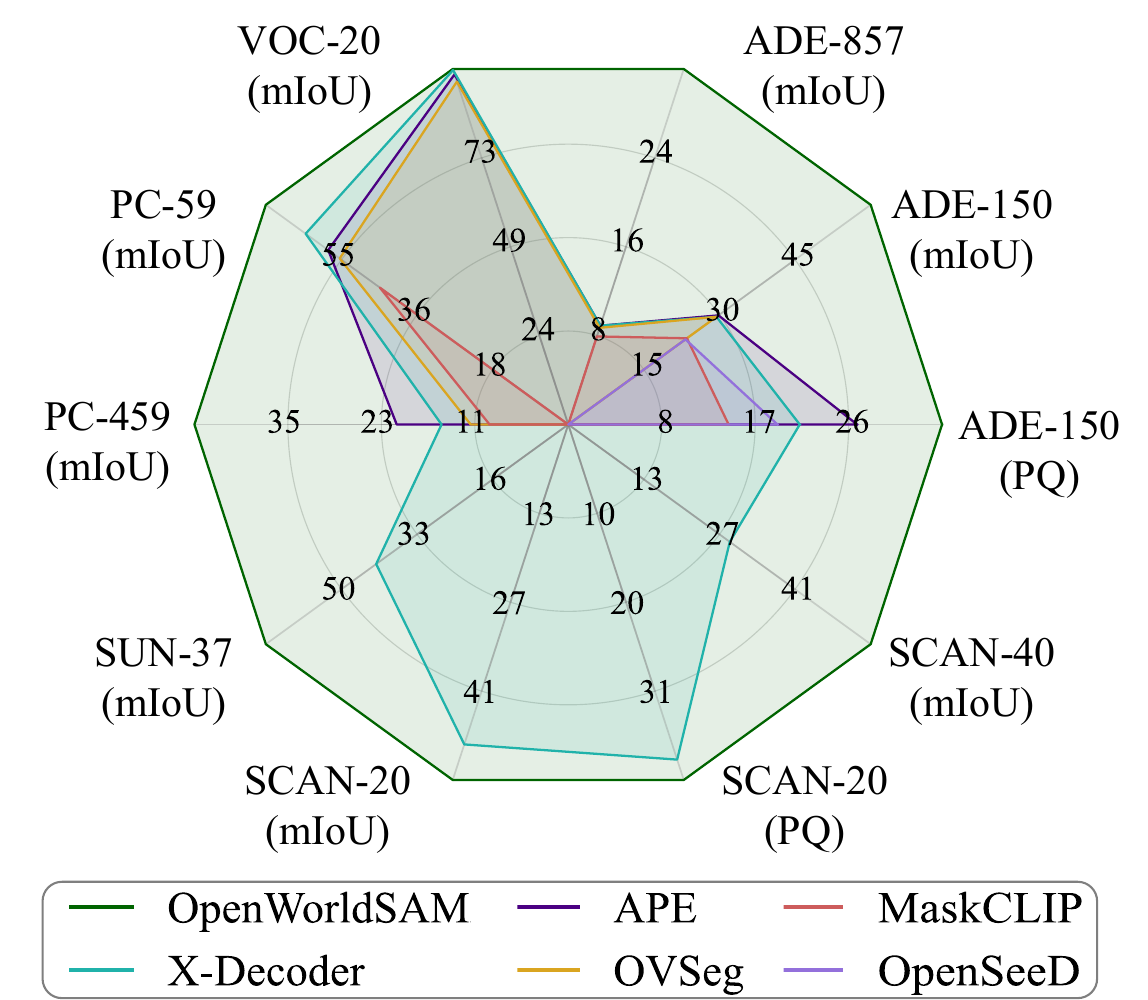}
  \end{center}
  \caption{OpenWorldSAM achieves new state-of-the-art  on six datasets with one suite of parameters.}
  \label{fig:radar}
\end{wrapfigure}

In summary, OpenWorldSAM represents a new paradigm of “\textit{segment anything in the open world}”. It inherits SAM’s interactiveness while being guided by flexible language prompts. Our contributions include:
\begin{enumerate}[leftmargin=*, nosep]
    \item We introduce OpenWorldSAM, a unified interface that supports various open-vocabulary segmentation tasks. We propose an efficient language adapter with tie‑breaker and cross‑attention soft prompting, improving multi-object localization.
    \item OpenWorldSAM achieves state-of-the-art zero-shot performance across six benchmarks (Figure~\ref{fig:radar}), setting a new standard for open-vocabulary segmentation (\eg, 60.4 mIoU on ADE20K~\cite{zhou2017ade20k}). OpenWorldSAM also acheives strong performance in referring expression segmentation (74.0 cIoU on RefCOCOg~\cite{nagaraja2016modeling}) with substantially fewer resources compared to recent models.
    \item Our work demonstrates that lightweight architectural interventions can unlock zero-shot segmentation capabilities rivaling specialized models while preserving SAM2’s efficiency and interactivity.
\end{enumerate}

\section{Related Work}
\textbf{Open-vocabulary segmentation.} Recent advances in open-vocabulary segmentation have leveraged vision-language models (VLMs)~\cite{radford2021clip, jia2021align} to overcome the constraints of traditional closed-set segmentation models. Early approaches like LSeg \cite{li2022lseg}, RegionCLIP~\cite{zhong2022regionclip} and OWL-ViT~\cite{minderer2022owlvit} established a baseline by introducing a contrastive learning framework to align image embeddings with CLIP-based text embeddings for zero-shot detection/segmentation. Subsequent methods~\cite{ghiasi2022openseg, xu2022groupvit} scaled effectively by using weak supervision of large-scale images with captions (up to millions of regions) or text-only signals, enabling more flexible and broader semantic coverage. Two-stage approaches like MaskCLIP \cite{dong2023maskclip} and OVSeg \cite{liang2023ovseg} further refined this paradigm by generating mask proposals using MaskFormer~\cite{cheng2021maskformer} followed by CLIP-based classification, notably boosting accuracy through mask-adapted fine-tuning. Another line of works formulated this task as a visual grounding problem and established region-text fusion~\cite{li2022glip, liu2024grounding, yan2023universal, yao2022detclip}. More recently, unified architectures such as ODISE~\cite{xu2023odise}, X-Decoder \cite{zou2023xdecoder}, SEEM \cite{zou2023seem}, OpenSeeD \cite{zhang2023openseed}, HIPIE~\cite{wang2023hipie}, Semantic-SAM~\cite{li2023semanticsam} and APE \cite{shen2024ape} have integrated multiple segmentation tasks into a single framework, showing significant progress towards general-purpose models, but they typically required resource intensive pre-training.

\textbf{Extending SAM for text-prompted segmentation.} The Segment Anything Model (SAM) \cite{kirillov2023sam, ravi2024sam2} achieved a breakthrough in promptable segmentation by training on 1 billion masks, enabling it to generate high-quality masks for visual prompts. A flurry of recent works have explored infusing SAM with semantic or language understanding to move beyond its original prompt types. Grounded-SAM~\cite{ren2024grounded-sam} is a pioneering effort that combines an open-vocabulary detector GroundingDINO~\cite{liu2024grounding} to generate bounding boxes from a text query, then feeds those boxes as prompts into SAM. FastSAM~\cite{zhao2023fastsam} matches CLIP embeddings with regions of interest. LLM-centric works~\cite{li2023refsam, rasheed2024glamm, lai2024lisa, xu2024ullava} attempt to map large LLMs or VLMs language embeddings into SAM or SAM-like decoder’s prompt latent space to enable referring expression segmentation. Among these, LISA~\cite{lai2024lisa} pioneered ``mask-as-text-embedding'' approach but was limited to single-object queries. LISA++~\cite{yang2023lisa2} introduced instance awareness through additional instruction-tuning data, though it requires LLMs to explicitly enumerate objects—a computationally expensive process.
EVF-SAM~\cite{zhang2024evfsam} recently demonstrated a lightweight alternative, integrating SAM with a multi-modal BEiT-3 encoder~\cite{wang2023beit} (0.7B parameters). While achieving state-of-the-art referring segmentation accuracy with minimal parameters, it remains constrained to single-object queries.
Inspired by the success of EVF-SAM, we enhance SAM further into the domain of open-vocabulary segmentation, where the goal is to segment and label all objects (``things'' and ``stuff'') in the scene with open-set categories.

\section{Methodology} \label{sec:method}

\begin{figure}[t]
    \centering
    \includegraphics[width=0.85\linewidth]{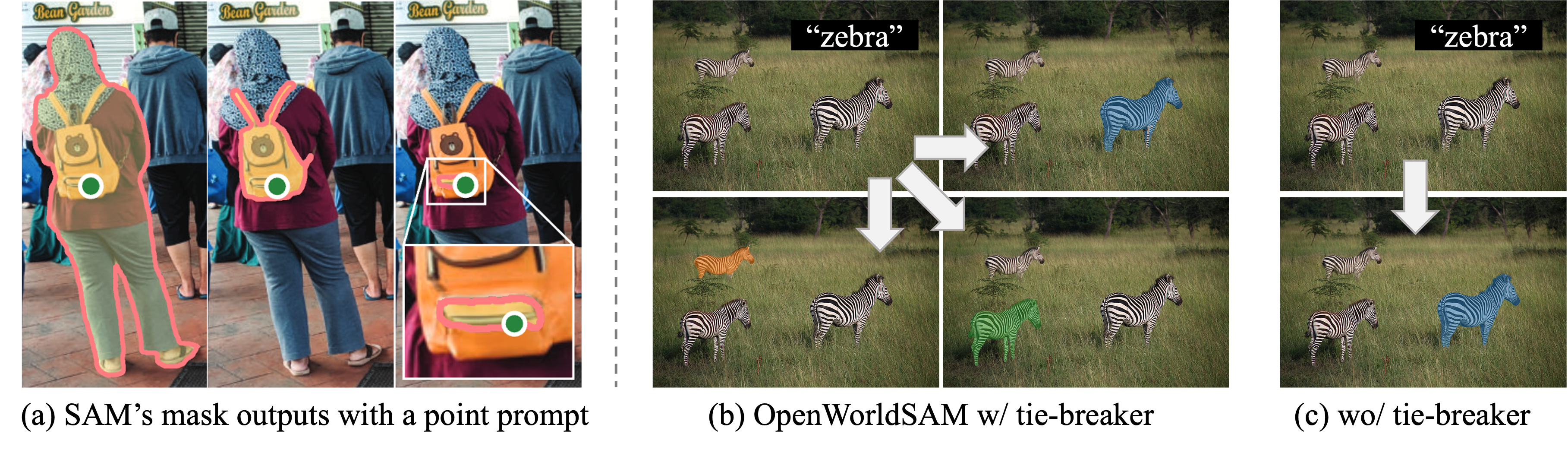}
    \caption{(a) SAM takes a visual click and outputs 3 valid masks on the same person (the person, the backpack, and a backpack region)~\cite{kirillov2023sam}. It will \textbf{not} output masks for the person standing next to her. (b) Tie-breakers shift the queries to distinct regions, enabling simultaneous segmentation of all three ``zebra'' instances. (c) Naïve approach~\cite{zhang2024evfsam}: A single language query for “zebra” causes SAM2 to segment only the most salient instance. }
    \label{fig:qualitative-zebra}
    \vspace{-15pt}
\end{figure}

\textbf{Motivation and key challenges.} A fundamental limitation of SAM-like architectures is their inability to resolve multi-instance ambiguity from a single prompt. While visual prompts (\eg, points) may occasionally lack granularity specificity—for instance, a click on a backpack could imply segmentation of either the backpack or the entire person (Figure~\ref{fig:qualitative-zebra}a)—they inherently localize to a single spatial region. Language prompts, however, introduce a distinct challenge: a text query like “zebra” may correspond to multiple spatially disjoint objects (Figure~\ref{fig:qualitative-zebra}b), with no prior knowledge of instance counts. Prior attempts to add language capabilities either rely on  segmentation-then-classification pipelines (losing end-to-end training) or require costly region-level text grounding during pre-training. 
Our key insight addresses this gap: SAM2’s mask decoder can inherently segment multiple instances if equipped with \textit{diverse positional guidance}, \ie, learned cues that disentangle identical semantic queries into spatially distinct segmentation targets.

\begin{figure}[t]
    \centering
    \includegraphics[width=0.95\linewidth]{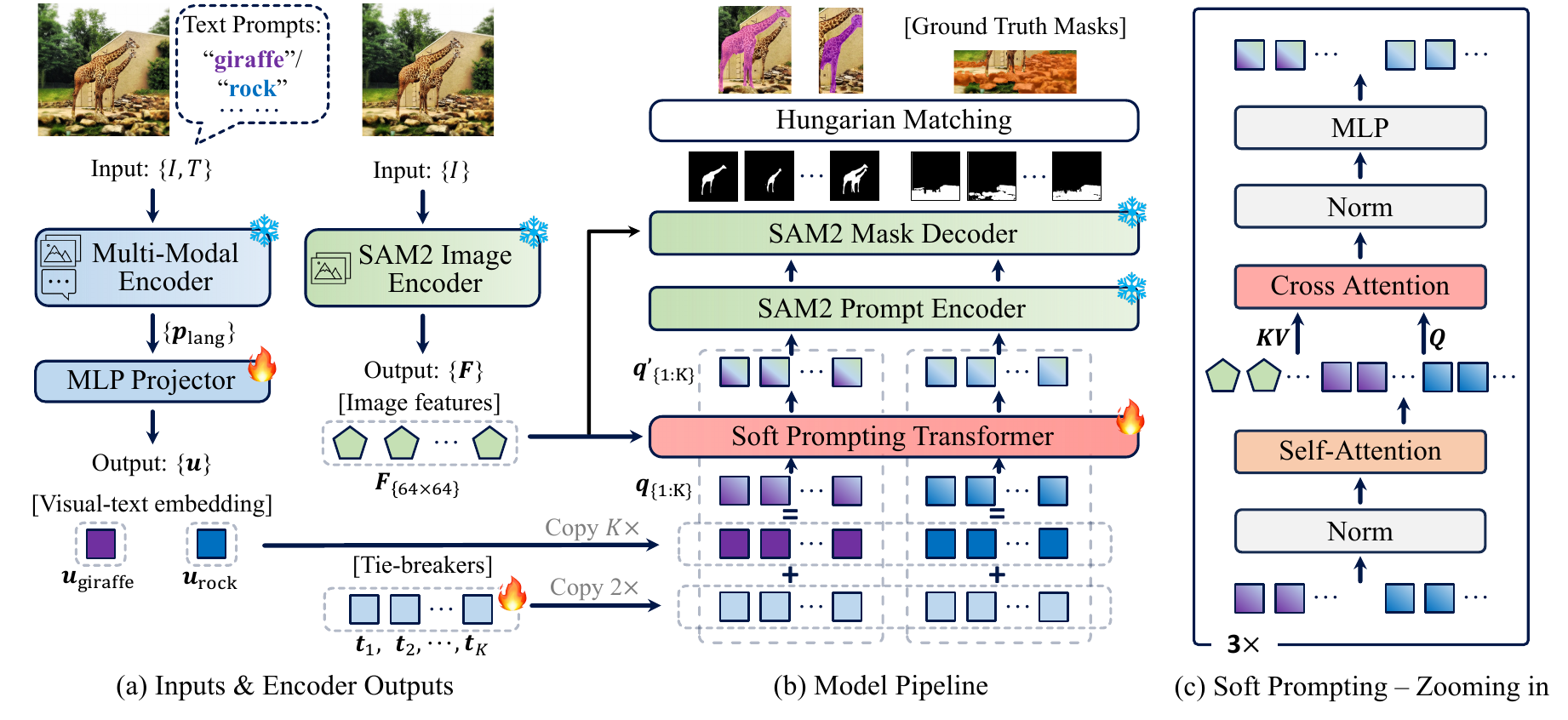}
    \caption{(a) Preliminaries on the inputs and outputs of the vision and multi-modal encoders. (b) OpenWorldSAM pipeline. (c) Detailed soft prompting Transformer architecture. }
    \label{fig:arch}
    \vspace{-10pt}
\end{figure}

\textbf{Architecture overview.} Figure \ref{fig:arch} depicts our framework which comprises: (i) a hierarchical SAM2 image encoder that extracts image features, (ii) a multi-modal vision‐language encoder that jointly ingests the image and text prompt, (iii) a lightweight MLP projector, (iv) learnable positional tie‐breakers for multi‐instance queries, (v) a soft prompting Transformer block that aligns text–image features with SAM2's image features, and (vi) the SAM2 mask decoder producing final masks. Only a small language adapter with components (iii–v) is trained; all other backbones remain frozen.

\textbf{Multi-modal encoder.}
We leverage BEiT-3~\cite{wang2023beit} to encode the input description into a semantic embedding. Given an image $I$ and a text prompt $T$ (e.g., a category name or a referring expression), we feed both modalities into BEiT‐3’s encoder to obtain joint visual–text embeddings.  Concretely, tokens of $T$ and patch embeddings of a downsampled $I$ are concatenated and processed by BEiT‐3, yielding a set of feature vectors $\{\mathbf{f}_\text{[CLS]}, \mathbf{f}_1, \dots\}$.  We take the classification token $\mathbf{f}_\text{[CLS]}$ as a compact summary denoted as $\mathbf{p}_\mathrm{lang}$ of the prompt conditioned on the image content. 

We adopt BEiT‑3 because its early‑fusion training on image‑text pairs equips it with rich, bidirectional semantics—crucial for reasoning about unseen classes. Compared with CLIP‑style contrastive image-text matching using only the features from the last encoder layers, BEiT‑3 exposes finer cross‑modal interactions. By embedding the text while it sees the image, the encoder already localizes the concept loosely (e.g., ``giraffe'' vs. ``rock'' in Figure~\ref{fig:arch}) before any downstream segmentation, preventing the mask decoder from learning semantics from scratch.

\textbf{Prompt projection.} BEiT‑3 emits 1,024‑D tokens, whereas SAM’s prompt channels are 256‑D.  
A two‑layer MLP acts as a projector that (i) preserves the coarse semantics of $\mathbf{p}_\mathrm{lang}\in\mathbb{R}^{d_{1024}}$ and (ii) learns to highlight dimensions that are most useful for mask prediction:
$
\mathbf{u} = \mathrm{MLP}(\mathbf{p}_\mathrm{lang})\in\mathbb{R}^{d_{256}}.
$

\textbf{Positional tie‑breaker and multi-instance queries generation.} 
The projected visual-text embedding $\mathbf{u}$ captures \textit{what} to segment but lacks awareness of \textit{how many} instances exist and \textit{where} they are. To enable multi-instance segmentation, we propose $K$  learnable positional tie-breaker vectors  $\{\mathbf{t}_1,\dots,\mathbf{t}_K\}\subset\mathbb{R}^{d_{256}}$ that perturb $\mathbf{u}$ into $K$ distinct queries:
\begin{equation}
    \mathbf{q}_i = \mathbf{u} + \mathbf{t}_i,\quad i=1,\dots,K.
\end{equation}
These perturbations serve two purposes: 1) Positional disambiguation: Each $\mathbf{t}_i$ nudges the query towards different spatial regions (Figure~\ref{fig:qualitative-zebra}b), mimicking how human annotators might click different points to segment each zebra. 2) Instance diversity: The tie-breakers are optimized during training to maximize coverage of distinct instances, preventing query collapse.
 Conceptually these queries play the role of the ``object queries'' in DETR~\cite{carion2020detr}. Crucially, they impose segmentation distinction for the same language semantics, making positional tie-breaking a novel and key feature for OpenWorldSAM. In practice $K=20$ covers $>$99\% images in COCO~\cite{lin2014coco}; for larger scenes $K$ can be increased trivially.

\textbf{Soft‐prompting via cross‐attention.}
The perturbed queries $\{\mathbf{q}_i\}$  interact with SAM2’s image features through a 3-layer Transformer~\cite{vaswani2017attention} in Figure~\ref{fig:arch}c, which alternates self‑attention (queries talk to each other, promoting diversity) and cross‑attention  (queries look at image features). Each language‑aware query
 is refined on‑the‑fly by cross‑attention with the frozen SAM2 features. SAM2’s image encoder follows a hierarchical vision Transformer (“Hiera”~\cite{ryali2023hiera, bolya2023window}) that outputs three features $\{\mathbf{F}_{256\times256}, \mathbf{F}_{128\times128}, \mathbf{F}_{64\times64} \}$ with $256^2$, $128^2$, and $64^2$ spatial resolutions, respectively. We operate on the level-3 features with $64^2$ resolution as they optimally balance precision for retainaing boundary details and computational efficiency ($16\times$ cheaper than full‑resolution attention). They are also used by SAM2 for mask decoding by default~\cite{ravi2024sam2}. The soft prompting Transformer computes
$
    \mathbf{q}'_i = \mathrm{CrossAttn}(\mathbf{q}_i,\ \mathbf{F}_{64\times64}),\; i=1,\dots,K,
$
whose key/value inputs are the flattened level-3 features $\mathbf{F}_{64\times64}\in\mathbb{R}^{4096\times256}$. This step grounds the language-aware queries in SAM2’s high-resolution visual features, resolving ambiguities (\eg, distinguishing adjacent zebras by stripe patterns). 

\textbf{Mask decoding and class assignment.}
The refined queries $\{\mathbf{q}'_i\}$ are input to SAM2’s mask decoder alongside level-3 image features. We inject the queries as the prompt tokens in place of, \eg, point or box prompts in the original SAM2's prompt encoder to obtain prompt embeddings. The prompt embeddings are then passed to the mask decoder which outputs $K$ masks and corresponding confidence scores.  We assign each mask the original text prompt $T$ as its class label, since the generation is fully conditioned on $T$ and thus inherits the semantic identity.

\textbf{Training.}
All heavy visual (Hiera) and vision‑language (BEiT‑3) encoders are kept frozen to preserve their pre‑trained knowledge and avoid costly retraining. Only the MLP projector, tie‑breakers, and the soft prompting Transformer are learnable. For each training sample and prompt, we match the $K$ predicted masks to the ground‐truth masks of class $T$ via Hungarian matching~\cite{carion2020detr}, then apply a focal loss, encouraging precise segmentation of all instances described by the prompt. The tie-breakers $\mathbf{t}_i \in \mathbb{R}^{d_{256}}$ are implemented as learnable parameters randomly initialized from a normal distribution. During training, the Hungarian matching loss naturally encourages each $\mathbf{t}_i$ to specialize in different spatial regions. Notably, this mechanism requires no explicit supervision about instance counts.

\textbf{Inference.} From the predicted $K$ masks, we derive results for three segmentation tasks: semantic, instance, and panoptic. For semantic segmentation, we merge masks sharing the same class label, weighted by their confidence scores. For instance segmentation, we apply confidence-score filtering to remove masks below a certain threshold, followed by non-maximum suppression (NMS) to eliminate highly overlapping masks and retain distinct object instances. Similarly, for panoptic segmentation, we perform confidence-based filtering and NMS, ensuring each pixel is uniquely assigned to either a ``thing'' (instance) or ``stuff'' (semantic) label.

Optionally, we perform a two-stage inference. In this setup, masks obtained from the first inference stage are used as visual prompts fed back into SAM2’s mask decoder, which refines mask contours. Qualitatively, two-stage inference improves the precision of mask boundaries for correct predictions (Appendix \ref{appendix:two-stage}). However, quantitative analysis (Table~\ref{tab:sota}) reveals that the second inference stage provides minimal improvements in segmentation metrics, suggesting it mainly enhances visual quality rather than overall accuracy.

\vspace{-8pt}

\section{Experiments} \label{sec:experiments}
 \textbf{Datasets and metrics.} 
 We train OpenWorldSAM on the COCO2017-Stuff~\cite{lin2014coco} dataset with panoptic annotations following X-Decoder~\cite{zou2023xdecoder}. The training set contains 104k images. We evaluate the model in a zero-shot setting on eight segmentation tasks across five diverse datasets: ADE20K-150/857~\cite{zhou2017ade20k}, PASCAL VOC-20~\cite{everingham2011pascalvoc}, PASCAL Context-59/459~\cite{mottaghi2014pascal-context}, ScanNet-20/40~\cite{dai2017scannet}, and SUN-RGBD-37~\cite{song2015sun}. Evaluation metrics include panoptic quality (PQ), mean average precision (mAP), and mean intersection-over-union (mIoU), corresponding to panoptic, instance, and semantic segmentation tasks, respectively. For referring segmentation, we  finetune on RefCOCOg UMD training split. Following prior works, we report the cumulative intersection over the cumulative union (cIoU) metric on the RefCOCOg UMD validation split.
 
 \textbf{Implementation.}  We implement our model in PyTorch.  We initialize the visual model with the  weights of SAM2-Hiera-Large~\cite{ravi2024sam2} and the VLM encoder with the weights of EVF-SAM BEIT-3-Large~\cite{zhang2024evfsam}. It is trained for 25 epochs on COCO-Stuff using the AdamW optimizer with a learning rate of 1e‑4, batch size 8, on a single NVIDIA A100 GPU. Image resolution is set to 1024 for SAM2 and 224 for BEiT-3. Number of postional tie-breaks is set to 20 for COCO dataset. Our implementation details can be found in Appendix \ref{appendix:experimental-settings}.

 \subsection{Open-Vocabulary Segmentation Evaluation Protocols and Challenges}

\textbf{Ambiguity of open vocabulary evaluation.}  Most prior open-vocabulary segmentation methods—including X-Decoder~\cite{zou2023xdecoder}, OVSeg~\cite{liang2023ovseg}, and MaskCLIP~\cite{dong2023maskclip}—adopt a \textbf{Global-Matching} protocol: for each predicted mask, a model matches it against the entire dataset vocabulary using precomputed text embeddings and selects the best-aligned class. However, this strategy can be problematic when applied to datasets like ADE20K, which contain hundreds of fine-grained and overlapping labels. As observed in OVSeg~\cite{liang2023ovseg}, this leads to semantically reasonable predictions being marked incorrect under exact label matching: \textit{“The ground-truth category is ‘building’ while our model predicts ‘skyscraper’. 
”} This ambiguity stems from the inherent subjectivity of language: synonymous or closely related concepts may be indistinguishable in a visual context, yet only one is accepted by the ground truth. We observe similar issues in our own qualitative analysis. As shown in Figure \ref{fig:qualitative-results}, X-Decoder predictions on ADE20K-857 often produce valid but non-canonical labels (e.g., ‘road’ instead of ‘runway’, or ‘screen’ instead of ‘arcade machine’), resulting in unfair penalization.

\textbf{Oracle-Prompts evaluation.} To address this, we introduce an alternative evaluation strategy: \textbf{Oracle Prompts}--during evaluation, we explicitly provide the ground-truth class names as prompts. This mimics the intended use case of \textbf{prompt-based models} like SAM, which are inherently interactive and conditioned on user input. Under this protocol, the model does not have to resolve linguistic ambiguity across the full label space; it segments \textit{what the user asks for}. We report results under both settings: Table \ref{tab:sota} shows baseline performance using the global matching protocol, consistent with prior works.
Table \ref{tab:oracle-results} revisits X-Decoder under the oracle-prompt protocol for a more equitable comparison to OpenWorldSAM, which by design is evaluated under oracle prompts.
We believe this approach provides a more fair assessment of SAM-style models in open-vocabulary segmentation.

\subsection{Open-Vocabulary Segmentation Performance Analysis}
\begin{figure}[t]
    \centering
    \includegraphics[width=\textwidth]{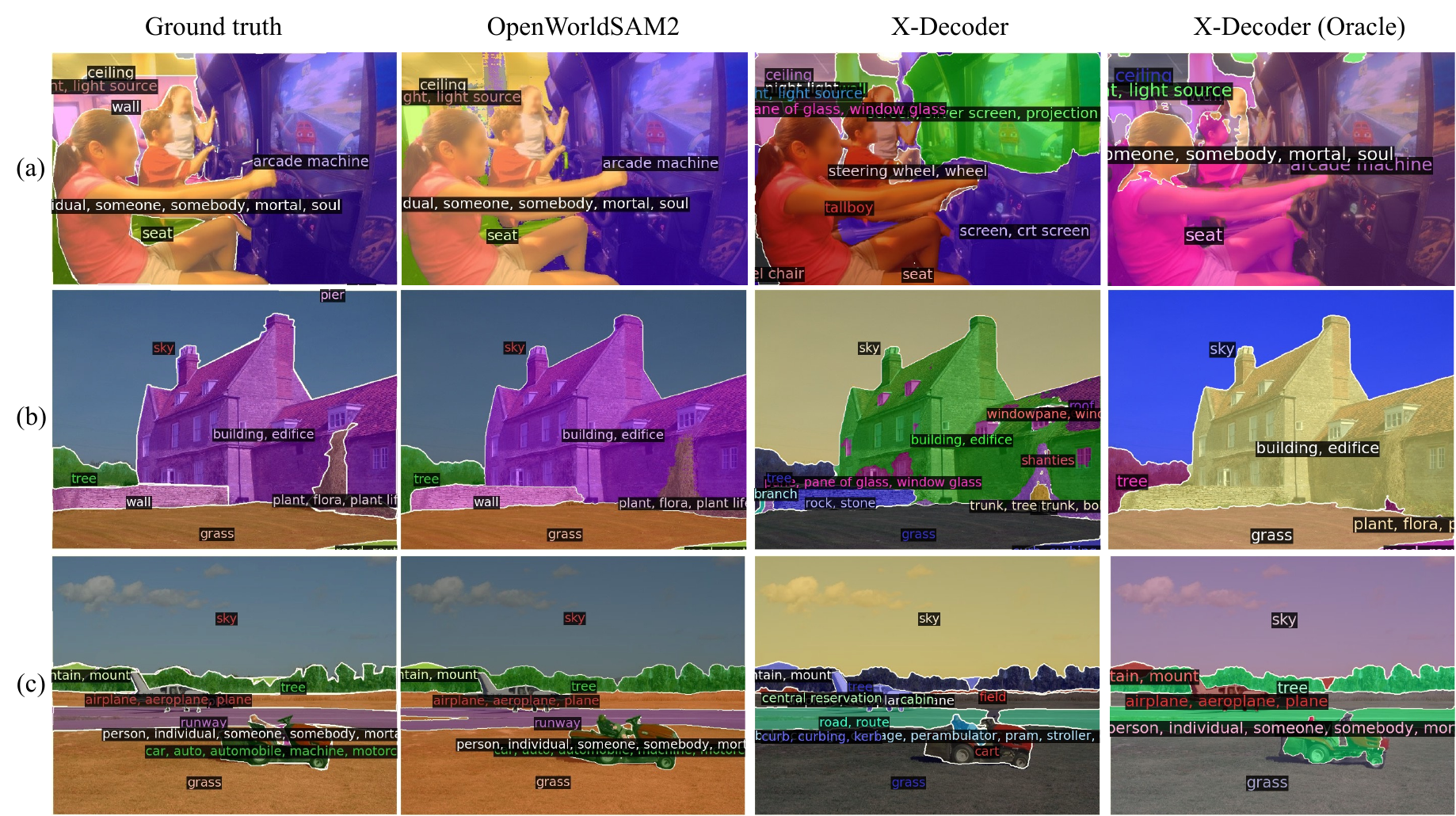}
    \caption{\textbf{Qualitative comparisons on ADE20K-857.} In many cases, (e.g., (c) road, field), X-Decoder predicts semantically related but incorrect labels due to ambiguity in the category list. The final column shows X-Decoder predictions using oracle prompts, which reduces confusion. OpenWorldSAM, conditioned on the correct prompt, produces faithful masks and avoids semantic mismatches. Color maps for each model vary. Please refer to the predicted labels. Best viewed with zoom in. We use two-stage inference for the  visualization. }
    \label{fig:qualitative-results}
    \vspace{-8pt}
\end{figure}

\begin{table}[t]
\centering
\scriptsize
\caption{\textbf{Zero-shot performance} of open-vocabulary segmentation models across multiple benchmarks. For COCO, different methods use different supervisions of mask (m), class label (cls) and caption (cap). “ITP” indicates whether model uses image-text pairs/referring data. ``DET” indicates extra detection data (\eg, Objects365, LVIS, OpenImages, \etc.) “*” denotes the model has the capability for the task but does not have number reported. “-” means the model does not have the ability for the specific task.  \textcolor{Orchid}{Purple} color means a fully supervised approach, and \textcolor{Apricot}{tan} a semi-supervised learning approach. Two-stage inference means we refine mask contours by re-prompting SAM using the raw mask predictions. Bold entries indicate the best performance. }
\renewcommand{\arraystretch}{1.2}
\setlength{\tabcolsep}{2pt}
\resizebox{\textwidth}{!}{\begin{tabular}{lcccccc|ccccccccccc}
\toprule
\multirow{2}{*}{Model} & \multirow{2}{*}{Train Params} & \multicolumn{3}{c}{COCO (p/s)} & \multirow{2}{*}{ITP} & \multirow{2}{*}{DET} &\multicolumn{3}{c}{ADE-150} & ADE-857 & VOC-20 & PC-59 & PC-459 & SUN-37 & \multicolumn{2}{c}{SCAN-20}  & SCAN-40  \\
& & m & cls & cap & & & PQ & mAP & mIoU & mIoU & mIoU & mIoU & mIoU & mIoU & mIoU & PQ & mIoU  \\
\midrule
MSeg (B)~\cite{lambert2020mseg}   &   70 (M)    & \cmark & \cmark & \cmark & \xmark & \xmark & \textcolor{Orchid}{33.7} & \textcolor{Orchid}{32.6} & 19.1 & * & 73.4 & 43.4 & * & 29.6 & 33.4 & * &  *  \\
GroupViT (S)~\cite{xu2022groupvit}    &  44 (M)       & \xmark & \xmark & \xmark & \cmark & \xmark & - & - & * & * & 52.3 & 22.4 & * & * & * & - &  *  \\
LSeg+ (B)~\cite{li2022lseg}    &  112 (M)     & \cmark & \cmark & \cmark & \xmark & \xmark & - &- & 18.0 & 3.8 & *  & 46.5 & 7.8 & * &  * & - & *  \\
ZegFormer (B)~\cite{ding2022zegformer}  & 60 (M)    & \cmark & \cmark & \cmark & \cmark & \xmark & - & - & * & \textcolor{Apricot}{8.1} & \textcolor{Apricot}{80.7} & * & * & * &  * & - & -  \\
OpenSeg (B)~\cite{ghiasi2022openseg}   &  86 (M)   & \cmark & \xmark & \cmark & \cmark & \xmark & - & - & 26.4 & 8.1 & 70.2 & 44.8 & 11.5 & * &  * & - & * \\
OVSeg (B)~\cite{liang2023ovseg} & 0.6  (M)  & \cmark & \cmark & \xmark & \xmark & \xmark & - & - & 29.6 & 9.0 &  94.5 &  55.7 & 12.4 & * & * & - & * \\
MaskCLIP (L)~\cite{dong2023maskclip}  &  428 (M)    & \cmark & \cmark & \xmark & \xmark & \xmark & 15.1 & 6.0 & 23.7 & 8.2 & * & 45.9 & 10.0 & * &  * & * & *  \\
OpenSeeD (L)~\cite{zhang2023openseed}  & 39 (M)    & \cmark & \xmark & \cmark & \cmark & \cmark & 19.7  & 15.0 & 23.4 & * & * & * & * & * &  * & * & *   \\
X-Decoder-Seg$^+$ (B)~\cite{zou2023xdecoder}  & 28 (M) & \cmark & \cmark & \xmark & \xmark & \xmark & 16.9 & 9.5 & 23.8 & 4.6 & 97.8 & 64.7 & 12.1 & 32.2 & 35.1 & 33.8 & 18.5  \\
X-Decoder (L)~\cite{zou2023xdecoder} & 38 (M)  & \cmark & \cmark & \cmark & \cmark & \xmark & {21.8} & {13.1} & {29.6} & {9.2} & 97.7 & 64.0 & {16.1} & {43.0} & {49.5} & {39.5} & 29.7  \\
APE-B (L)~\cite{shen2024ape} & 42 (M) & \cmark & \cmark & \cmark & \cmark & \cmark & 26.4 & \textbf{23.5} & 29.0 & 9.2 & 95.8 & 58.3 & 21.0 & * &  * & * & *  \\
ESC-Net~\cite{lee2025effective} & 451 (M) &  \cmark & \cmark & \xmark & \xmark & \xmark & - & - & 41.8 & 18.1 & \textbf{98.3} & 65.6 & 27.0 &  * &  * & - & *  \\
\midrule
\rowcolor{gray!15}
OpenWorldSAM & {4.5 (M)} & \cmark & \cmark & \xmark & \cmark & \xmark & {35.2} & 16.9 &  \textbf{60.4} &   \textbf{33.1}  &  {98.0}  &   \textbf{73.7}  &   \textbf{47.5}  &   {67.7}  &   \textbf{65.0}  &  \textbf{41.9}  &   \textbf{55.6} \\
\rowcolor{gray!15}
\; + two-stage inference  & 4.5 (M) & \cmark & \cmark & \xmark & \cmark & \xmark & \textbf{36.3} & 15.6 &  58.0 &   32.6  &  97.6  &   72.6  &   45.8  &   \textbf{68.2}  &   64.8  &  39.9  &   54.1 \\
\bottomrule
\end{tabular}}
\label{tab:sota}
\vspace{-12pt}
\end{table}

\begin{table}[h]
\centering
\scriptsize
\caption{\textbf{Oracle-Prompts evaluation} of open-vocabulary segmentation models. We report the state-of-the-art (SOTA) model X-Decoder~\cite{zou2023xdecoder}'s performance under both evaluation protocols. Other methods are omitted either because: 1) they are not SOTA, or 2) they do not support oracle-prompts evaluation.}
\renewcommand{\arraystretch}{1.2}
\setlength{\tabcolsep}{2pt}
\begin{tabular}{llccccccccccc}
\toprule
\multirow{2}{*}{Model} &  \multirow{2}{*}{Evaluation Protocol} &ADE-150 & ADE-857 & VOC-20 & PC-59 & PC-459 & SUN-37 &  SCAN-40  \\
&  & mIoU & mIoU & mIoU & mIoU & mIoU & mIoU  & mIoU  \\
\midrule
X-Decoder (L)~\cite{zou2023xdecoder}    & Global-Matching (default) \quad  & {29.6} & {9.2} & 97.7 & 64.0 & {16.1} & {43.0}  & 29.7  \\
X-Decoder (L)          & Oracle-Prompts  & 51.5 & 29.1 & \textbf{98.1}  & \textbf{75.5} & 42.3 &  67.1  &  49.1 \\
\rowcolor{gray!15}
OpenWorldSAM \quad\quad  & Oracle-Prompts (default)\qquad  &  \textbf{60.4} &   \textbf{33.1}  &  {98.0}  &   {73.7}  &   \textbf{47.5}  &   \textbf{67.7}  &    \textbf{55.6} \\
\bottomrule
\end{tabular}
\label{tab:oracle-results}
\vspace{-12pt}
\end{table}

\textbf{Zero-shot open-vocabulary transfer.} OpenWorldSAM generalizes out-of-the-box to a broad set of segmentation tasks without any weight adaptation. As shown in Table \ref{tab:sota}, it achieves state-of-the-art performance across almost all datasets and evaluation metrics. Its performance consistently surpasses strong baselines such as X-Decoder and APE, despite using only 4.5M trainable parameters.  On ADE20K-857, OpenWorldSAM achieves 33.1\% mIoU, outperforming the previous best (X-Decoder) by +23.9 absolute points (9.2 → 33.1). On PASCAL Context-459, it achieves 47.5\% mIoU, improving over APE's 21.8\% by +25.7 points, and on ScanNet-40, it reaches 55.6\% mIoU, a +25.9 point improvement over X-Decoder's 29.7\%. On AP score we under-perform APE, which included extra detection datasets, \eg, Objects365~\cite{shao2019objects365}, in their training recipe for better localization. 

We attribute our strong performance to the model’s prompt-conditioned decoding mechanism, which directly leverages language input to guide mask prediction. This is particularly advantageous when the target concept is known at query time. In contrast, global retrieval-based models such as X-Decoder must resolve ambiguity across the entire vocabulary space, which introduces classification error. While one might argue that differing evaluation protocols confound the comparison, it’s important to note that both families of models require the same semantic input—the only difference lies in \textbf{when} and \textbf{how} that input is used.

\textbf{Oracle-Prompts evaluation.} As SAM-style models are designed for interactive segmentation, oracle prompts closely reflect practical use cases—such as human-in-the-loop annotation, robotic object search, or dynamic UI feedback. To fairly compare with the state-of-the-art generalist model X-Decoder~\cite{zou2023xdecoder}, we also evaluate it under oracle prompts: we restrict its vocabulary to the ground-truth classes for each image. As shown in Table \ref{tab:oracle-results}, OpenWorldSAM continues to outperform even under these controlled conditions. Notably, on large-vocabulary datasets such as ADE20K-857 and PASCAL Context-459, OpenWorldSAM achieves 33.1\% and 47.5\% mIoU, surpassing X-Decoder by +4.0 and +5.2 points, respectively. This highlights our model’s superior language grounding ability in long-tailed, fine-grained category distributions. On smaller datasets like PASCAL Context-59 and PASCAL VOC-20, where most categories overlap with COCO, X-Decoder slightly outperforms our model (75.5\% vs. 73.7\% mIoU and 98.1\% vs. 98.0\%), suggesting it benefits more from class memorization in such settings. Moreover, Figure \ref{fig:qualitative-results} illustrates that global matching often fails despite producing correct masks. Conditioning on oracle prompts significantly reduces this ambiguity, highlighting the robustness of our evaluation protocol and the effectiveness of prompt-based segmentation. 

\textbf{Qualitative Results.}  Figure \ref{fig:qualitative-results} presents example outputs of OpenWorldSAM on challenging scenes, with comparisons to X-Decoder under both evaluation protocols. In one example \ref{fig:qualitative-results}(a), an image from ADE20K-857 containing a game room scene is segmented by our model using prompts for various objects (``ceiling, light, seat, person, arcade machine''). OpenWorldSAM accurately masks each object and stuff region, whereas X-Decoder misclassifies the ``arcade machine'' due to confusion between similar semantic objects under Global-Matching, and produces fragmented masks for the person and seat under Oracle-Prompts. Similarly in example \ref{fig:qualitative-results}(b), X-Decoder misclassifies the ``wall'' and proposes object masks for prompts that did not exist in the ground truth (\eg, ``window glass'') under Global-Matching, and failed to segment ``plant'' under Oracle-Prompts. This showcases our model’s clear understanding of category semantics (thanks to the VLM prompt) combined with precise mask delineation (thanks to SAM2’s capability). More qualitative results in Appendix \ref{appendix:qualitative}.

\subsection{Referring Expression Segmentation Performance Analysis}

\begin{figure}[t]
    \centering
    \includegraphics[width=0.95\linewidth]{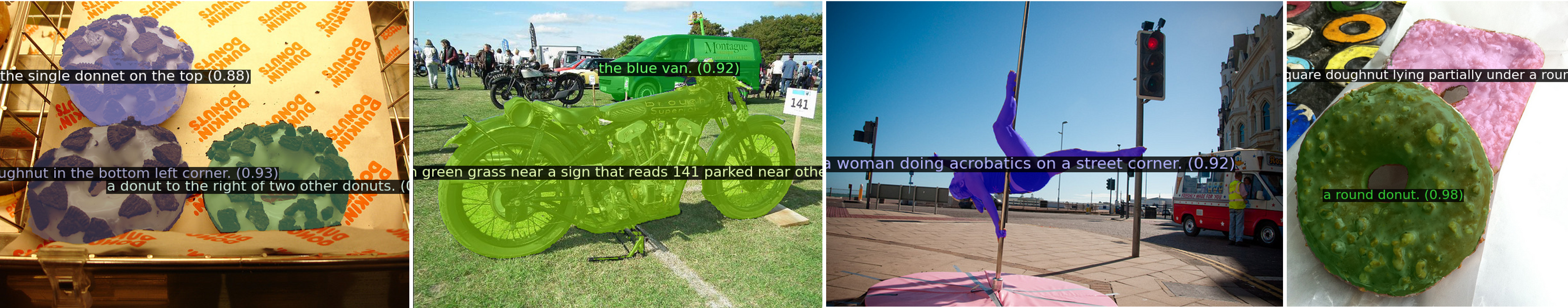}
    \caption{\textbf{Qualitative results on RefCOCOg.} OpenWorldSAM is capable of understanding spatial relationship, colors, actions, and shapes, \etc.}
    \label{fig:qualitative-referring}
    \vspace{-15pt}
\end{figure}

\begin{table}[t]
\centering
\scriptsize
\caption{\textbf{Referring segmentation performance (cIoU) comparison on RefCOCOg benchmark validation set} between our proposed OpenWorldSAM and prior SOTA methods. We abbreviate the datasets: C (COCO),  RC (RefCOCO/+), RCg (RefCOCOg), PL (PACO-LVIS), O365 (Objects365), V (Video segmentation datasets), OID (OpenImages Detection), VG (Visual Genome), ADE (ADE20K), PP (PASCAL-Part), PC (PASCAL-VOC). We compare model trainable parameters, model capabilities (OV seg (open-vocabulary segmentation) and Inter Seg (interactive segmentation)), and training data required. ``*'' denotes an estimate of the trainable parameters, since these models use LoRA~\cite{hu2022lora} with rank-8/16 adapters for finetuning.}
\renewcommand{\arraystretch}{1.2}
\setlength{\tabcolsep}{2.5pt}
\resizebox{\textwidth}{!}{\begin{tabular}{lllccclc}
\toprule
Method & Foundation Model & Train Params & w/ SAM? & OV Seg? & Inter Seg? &  {Training Data}  & {cIoU} \\
\midrule
X-Decoder (L)~\cite{zou2023xdecoder}     & CLIP-B~\cite{radford2021clip} (63M)      & \quad 38 (M)   & \xmark & \cmark & \xmark & C, RCg, Cap4M & 64.6 \\
SEEM (L)~\cite{zou2023seem} & CLIP-B~\cite{radford2021clip} (63M) & \quad 39 (M) & \xmark & \cmark & \cmark & C, RC, RCg, PL & 65.6 \\
PolyFormer (L)~\cite{liu2023polyformer}\quad  & BERT-B~\cite{devlin2019bert} (104M)    & \quad 450 (M)   & \xmark & \xmark & \xmark & RC, RCg    & 71.2 \\
UNINEXT (H)~\cite{yan2023universal}       & BERT-B~\cite{devlin2019bert} (104M)     & \quad 673 (M)  & \xmark & \cmark & \xmark &C, RC, O365, V & 74.4 \\
APE-B (L)~\cite{shen2024ape}         & CLIP-L~\cite{radford2021clip} (123M)    & \quad 42 (M)    & \xmark & \cmark & \xmark &C, PC, O365, OID, VG, RC, RCg \qquad & 63.5 \\
PixelLM~\cite{ren2024pixellm}           & LLaMA2~\cite{touvron2023llama2} (13B)   &   \quad 29 (M)$^*$   & \xmark & \cmark & \xmark &C, RC, ADE,  PL, MUSE & 69.3 \\
\midrule
LISA~\cite{lai2024lisa}              & Vicuna~\cite{vicuna2023} (7B)      & \quad 32 (M)$^*$   & \cmark& \xmark & \xmark &C, RC, ADE,  PL, PP & 66.4 \\
GLaMM~\cite{rasheed2024glamm}             & Vicuna~\cite{vicuna2023} (7B)     & \quad 40 (M)$^*$     & \cmark& \cmark & \xmark &RC, GranD & 74.2 \\
u-LLaVA~\cite{xu2024ullava}           & Vicuna~\cite{vicuna2023} (7B)     &  \quad 44 (M)$^*$   & \cmark& \xmark & \xmark &C, RC, ADE,  PL, PC & 71.6 \\
u-LLaVA~\cite{xu2024ullava}           & Vicuna~\cite{vicuna2023} (7B)     & \quad 7 (B)    & \cmark& \xmark & \xmark &C, RC, ADE,  PL, PC & 74.8 \\
Sa2VA~\cite{yuan2025sa2va}           & InternVL2~\cite{chen2024far} (1B)  & \quad 22 (M)$^*$ & \cmark & \xmark & \cmark & RC, RCg, V, GranD & 72.3 \\
Sa2VA~\cite{yuan2025sa2va}           & InternVL2~\cite{chen2024far} (4B)  & \quad 35 (M)$^*$ & \cmark & \xmark & \cmark & RC, RCg, V, GranD & 74.1 \\
SAMWISE~\cite{cuttano2025samwise} & RoBERTa~\cite{liu2019roberta} (125M) & \quad 5.6 (M) & \cmark& \xmark & \xmark & RC & 67.3 \\
EVF-SAM~\cite{zhang2024evfsam}         & BEIT-3-L~\cite{wang2023beit} (0.7B)  & \quad 674 (M)   & \cmark& \xmark & \xmark &RC & \textbf{77.0} \\
\rowcolor{gray!15}
OpenWorldSAM    & BEIT-3-L~\cite{wang2023beit} (0.7B)   &  \quad 4.5 (M)   & \cmark& \cmark & \cmark &C, RCg & 74.0 \\
\bottomrule
\end{tabular}}
\label{tab:referring-results}
\vspace{-16pt}
\end{table}

\textbf{Performance.} As shown in Table~\ref{tab:referring-results} and Figure~\ref{fig:qualitative-referring}, OpenWorldSAM achieves strong performance on the RefCOCOg validation set, obtaining a cIoU of 74.0\%, significantly outperforming earlier generalist models like SEEM and X-Decoder ($\approx$65\%), and competitive with specialized models such as GLaMM (71.2\%) and UNINEXT (74.4\%). Notably, OpenWorldSAM reaches this accuracy using just BEiT-3 encoder with 673M parameters and an additional 4.5M trainable parameters, substantially fewer than recent large-scale models like LISA, GLaMM, and u-LLaVA, which rely on much larger vision-language foundations (7B+ parameters) and multiple additional datasets. While EVF-SAM achieves higher cIoU (77.0\%), OpenWorldSAM inherits SAM’s interactive features, offering unique flexibility in interative segmentation tasks. Furthermore, with its tie-breaker and soft-prompting modules, OpenWorldSAM can also perform general segmentation tasks, distinguishing it from higher-scoring yet narrower models.

\begin{table}[t]
\centering
\scriptsize
\caption{\textbf{Ablation on the VLM choice,} \eg, CLIP~\cite{radford2021clip} model from OpenAI. ‘\cmark’ denotes modality used during training, and ‘\xmark’ unused. Only the adapter modules are trainable, and the VLMs are kept frozen. Late fusion means we concatenate text/image features from the last layers of CLIP's text/image encoder. Early fusion means BEiT-3 processes both modalities in all 24 Transformer layers.}
\renewcommand{\arraystretch}{1.2}
\setlength{\tabcolsep}{3pt}
\begin{tabular}{llcccccccc}
\toprule
\multirow{2}{*}{Encoder} &  \multirow{2}{*}{Params} & \multirow{2}{*}{Text} & \multirow{2}{*}{Image} & \multirow{2}{*}{Modality Fusion} &  \multicolumn{3}{c}{ADE-150}  &  ADE-857 & RefCOCOg \\
& & & & & PQ & AP & mIoU & mIoU & cIoU \\
\midrule
CLIP-Large & 123 (M) &  \cmark & \xmark & -- & 13.5 & 2.9 & 25.7 & 12.8 & 25.2 \\
CLIP-Large & 428 (M)& \cmark & \cmark  & Late (Last-layer Concat)   & 14.0 & 3.6 & 26.5  & 14.0 & 25.3\\
BEiT-3-Large \quad & 370 (M) & \cmark & \xmark  & -- & 13.6 & 3.1 & 26.3  & 13.3 & 26.1\\
\rowcolor{gray!15}
BEiT-3-Large & 673 (M) & \cmark & \cmark  & Early (All-layer Attention)  &  \textbf{35.2} & \textbf{16.9} &  \textbf{60.4} &   \textbf{33.1} & \textbf{74.0}\\
\bottomrule
\end{tabular}
\label{tab:ablation-vlm}
\vspace{-10pt}
\end{table}

\begin{table}[t]
    \centering
    \scriptsize
    \captionof{table}{\textbf{Ablation on trainable and inference modules. } For training, ‘\cmark’ denotes trainable, and ‘\xmark’ denotes frozen. For inference, ‘\cmark’ denotes activate, and ‘\xmark’ denotes non-activate.}
    \renewcommand{\arraystretch}{1.2}
\setlength{\tabcolsep}{2.6pt}
    \begin{tabular}{l|ccccl|ccc|cccc}
        \toprule
        \multirow{2}{*}{Exp} & \multicolumn{4}{c}{Train Modules}& \multirow{2}{*}{Train Params} & \multicolumn{3}{c}{Inference Modules} & \multicolumn{3}{c}{ADE-150} & ADE-857  \\
        & Tie-breaker & BEiT-3 & Cross-Attn & MLP Projector & &Tie-breaker & BEiT-3 & Cross-Attn & PQ & AP & mIoU & mIoU \\
        \midrule
        E1 & \xmark & \xmark & \xmark & \cmark & 1.2 (M) & \xmark & \cmark & \xmark & 0.4 & 1.0 & 1.2 & 0.2\\
        E2 &\cmark & \xmark & \xmark & \cmark & 1.3 (M) & \cmark & \xmark & \xmark & - & 9.5 & - & - \\
        E3 &\cmark & \xmark & \xmark & \cmark & 1.3 (M) & \cmark & \cmark & \xmark & 35.1 & \textbf{17.1} & 56.8 & 32.2 \\
        E4 &\cmark & \cmark & \xmark & \cmark & 674.0 (M) & \cmark & \cmark & \xmark & 13.6 & 3.5 & 24.4 & 10.6 \\
        \rowcolor{gray!15}
        E5 &\cmark & \xmark & \cmark & \cmark & 4.5 (M) & \cmark & \cmark & \cmark & \textbf{35.2} & 16.9 & \textbf{60.4} & \textbf{33.1} \\
        E6 & \cmark & \cmark & \cmark & \cmark & 677.2 (M) & \cmark & \cmark & \cmark & 15.9 & 3.8 & 23.6 & 11.2 \\
        \bottomrule
        \end{tabular}
	\label{tab:ablation-training-modules}
    \vspace{-10pt}
\end{table}

\subsection{Ablation Studies}

We systematically validate OpenWorldSAM’s design through zero-shot transfer on ADE20K-150/857 benchmark and fintuning on RefCOCOg benchmark.

\textbf{Multi-modal encoder analysis.} In Table~\ref{tab:ablation-vlm}, we compares performances using different VLM encoders and fusion methods (early fusion vs. late fusion). BEiT-3’s early cross-modal fusion (joint text-image processing across all layers) outperforms CLIP’s late fusion (last-layer concatenation) by +33.9 mIoU, +21.2 PQ, and +13.3 AP on ADE-150, demonstrating that deep semantic integration is critical for aligning language concepts with visual regions, echoing findings by EVF-SAM~\cite{zhang2024evfsam}.

\textbf{Visual Context Matters.} Table~\ref{tab:ablation-vlm} demonstrates that removing visual inputs to BEiT-3 (text-only) causes catastrophic performance collapse (-34.4 mIoU on ADE-150). This confirms that SAM’s segmentation backbone cannot ground textual semantics without explicit visual-textual co-encoding.



\textbf{Optimal Training Strategy.} In Table~\ref{tab:ablation-training-modules}, we varied the trainable modules in OpenWorldSAM (thus varying total new parameters from ~1.2M to ~770M). We found in E5 that freezing  BEiT-3 and training only the language adapter module (tie-breaker + cross-attention, 4.5M parameters) yields optimal performance (60.4 mIoU ADE-150). Notably, comparing E6 vs E5 and E4 vs E3, we found fine-tuning the entire BEiT-3 encoder (673M parameters) significantly degrades accuracy (mIoU drops from 60.4 to 23.6), likely due to underfitting on sparse category label prompts compared to its original web-scale pretraining.

\textbf{Positional tie-breaker vs. none. } Comparing E3 vs E1 in Table~\ref{tab:ablation-training-modules}, positional tie-breaker boosts AP from 1.0\% to 17.1\%.  As shown in Figure~\ref{fig:qualitative-zebra}, without the tie-breaker, the model usually collapses on one instance of the class (especially if the one was particularly salient among others). This confirms the necessity of this component for reliable instance segmentation.

\textbf{Cross-Attention layer removal.} As shown in Table~\ref{tab:ablation-training-modules}, E5 vs E3, removing the cross-attention layers expectedly led to inferior performance (-1.5 mIoU on ADE-150 and -0.9 mIoU on ADE-857). This indicates that cross-attention helps align prompts to the intended visual regions.

\section{Conclusion}
OpenWorldSAM bridges the gap between promptable segmentation and open-vocabulary understanding by unifying SAM’s segmentation prowess with vision-language models’ semantic grounding. This approach generalizes across tasks (semantic/instance/panoptic) and prompts (nouns/sentences), offering practitioners a unified tool for real-world scenarios where novel objects and ambiguous queries are the norm. Three innovations drive this success: (1) Positional tie-breakers enable multi-instance segmentation from single-text queries, resolving a critical limitation of SAM-like architectures. (2) Cross-modal soft prompting dynamically aligns language semantics with SAM’s visual space, ensuring precise localization without costly LLMs. (3) Frozen foundation synergy leverages pre-trained knowledge from SAM and BEiT-3, proving that dense prediction tasks benefit as much as classification from parameter-efficient adaptation. Beyond technical contributions, OpenWorldSAM advances a paradigm for extending segmentation foundations: instead of training monolithic models, strategic adaptation of frozen components achieves open-world readiness at minimal cost.

\textbf{Acknowledgement.} This work was supported in part by CoCoSys, a JUMP2.0 center sponsored by DARPA and SRC,
the National Science Foundation (CAREER Award, Grant \#2312366, Grant \#2318152), the DARPA Young Faculty Award and the DoE MMICC center SEA-CROGS (Award \#DE-SC0023198).

\bibliographystyle{unsrt}
\bibliography{references}

\newpage


\newpage
\section*{NeurIPS Paper Checklist}


\begin{enumerate}

\item {\bf Claims}
    \item[] Question: Do the main claims made in the abstract and introduction accurately reflect the paper's contributions and scope?
    \item[] Answer: \answerYes{} 
    \item[] Justification: In our abstract and introduction, we talk about our contribution on proposing a novel framework for open-vocabulary segmentation. We provide extensive experiments on comprehensive datasets to support this claim. We also conduct in-depth ablation studies that verifies the effectiveness of our model design.

\item {\bf Limitations}
    \item[] Question: Does the paper discuss the limitations of the work performed by the authors?
    \item[] Answer: \answerYes{}{} 
    \item[] Justification: We discuss the limitations in Appendix \ref{appendix:limitation}, which is about the model generalization quality to outdoor scenes and self-driving scenes.

\item {\bf Theory assumptions and proofs}
    \item[] Question: For each theoretical result, does the paper provide the full set of assumptions and a complete (and correct) proof?
    \item[] Answer: \answerNA{} 
    \item[] Justification: Our work does not include theoretical assumptions and proofs.

    \item {\bf Experimental result reproducibility}
    \item[] Question: Does the paper fully disclose all the information needed to reproduce the main experimental results of the paper to the extent that it affects the main claims and/or conclusions of the paper (regardless of whether the code and data are provided or not)?
    \item[] Answer: \answerYes{}{} 
    \item[] Justification: We provide the detailed methodology and experimental setup in Sections \ref{sec:method} and \ref{sec:experiments}. Moreover, we provide all source codes to reproduce the results, including training scripts (detailed configurations included) and evaluation scripts (model checkpoints included). We will open source the code on GitHub after acceptance.

\item {\bf Open access to data and code}
    \item[] Question: Does the paper provide open access to the data and code, with sufficient instructions to faithfully reproduce the main experimental results, as described in supplemental material?
    \item[] Answer: \answerYes{} 
    \item[] Justification: We provide the source codes as supplementary material. We provide instructions that contain the exact command and environment needed to run to reproduce the results. We provide instructions on data access and preparation, including how to access the raw data, preprocessed data, intermediate data, and generated data, etc.

\item {\bf Experimental setting/details}
    \item[] Question: Does the paper specify all the training and test details (e.g., data splits, hyperparameters, how they were chosen, type of optimizer, etc.) necessary to understand the results?
    \item[] Answer: \answerYes{} 
    \item[] Justification: Justification: Training and test details can be found in Sections \ref{sec:method} and \ref{sec:experiments}, and Appendix A.

\item {\bf Experiment statistical significance}
    \item[] Question: Does the paper report error bars suitably and correctly defined or other appropriate information about the statistical significance of the experiments?
    \item[] Answer: \answerNo{} 
    \item[] Justification: We did not include the error bars as we fix the random seed for every experiment, reducing the impact from data loader and other parameters initialization.

\item {\bf Experiments compute resources}
    \item[] Question: For each experiment, does the paper provide sufficient information on the computer resources (type of compute workers, memory, time of execution) needed to reproduce the experiments?
    \item[] Answer: \answerYes{} 
    \item[] Justification: We provide the details of computer resources in Section \ref{sec:experiments}. All experiments can be run on a single A100 GPU. We also provide analysis on trainable parameters in Section \ref{sec:experiments}.
    
\item {\bf Code of ethics}
    \item[] Question: Does the research conducted in the paper conform, in every respect, with the NeurIPS Code of Ethics \url{https://neurips.cc/public/EthicsGuidelines}?
    \item[] Answer: \answerYes{} 
    \item[] Justification: Our experiments conform to the NeurIPS Code of Ethics.

\item {\bf Broader impacts}
    \item[] Question: Does the paper discuss both potential positive societal impacts and negative societal impacts of the work performed?
    \item[] Answer: \answerNA{} 
    \item[] Justification: There is no social impact of this work.

\item {\bf Safeguards}
    \item[] Question: Does the paper describe safeguards that have been put in place for responsible release of data or models that have a high risk for misuse (e.g., pretrained language models, image generators, or scraped datasets)?
    \item[] Answer: \answerNA{} 
    \item[] Justification: The paper poses no such risks.

\item {\bf Licenses for existing assets}
    \item[] Question: Are the creators or original owners of assets (e.g., code, data, models), used in the paper, properly credited and are the license and terms of use explicitly mentioned and properly respected?
    \item[] Answer: \answerYes{} 
    \item[] Justification: Our model and its code development are based on baseline works which are credited in
the paper. Our datasets are the standard benchmarks that are widely used in academia.

\item {\bf New assets}
    \item[] Question: Are new assets introduced in the paper well documented and is the documentation provided alongside the assets?
    \item[] Answer: \answerNA{} 
    \item[] Justification: The paper does not release new assets.

\item {\bf Crowdsourcing and research with human subjects}
    \item[] Question: For crowdsourcing experiments and research with human subjects, does the paper include the full text of instructions given to participants and screenshots, if applicable, as well as details about compensation (if any)? 
    \item[] Answer: \answerNA{} 
    \item[] Justification: The paper does not involve crowdsourcing nor research with human subjects.

\item {\bf Institutional review board (IRB) approvals or equivalent for research with human subjects}
    \item[] Question: Does the paper describe potential risks incurred by study participants, whether such risks were disclosed to the subjects, and whether Institutional Review Board (IRB) approvals (or an equivalent approval/review based on the requirements of your country or institution) were obtained?
    \item[] Answer: \answerNA{} 
    \item[] Justification: The paper does not involve crowdsourcing nor research with human subjects.

\item {\bf Declaration of LLM usage}
    \item[] Question: Does the paper describe the usage of LLMs if it is an important, original, or non-standard component of the core methods in this research? Note that if the LLM is used only for writing, editing, or formatting purposes and does not impact the core methodology, scientific rigorousness, or originality of the research, declaration is not required.
    \item[] Answer: \answerNA{} 
    \item[] Justification: The core method development in this research does not involve LLMs as any important, original, or non-standard components.

\end{enumerate}

\newpage

\appendix

\section{Experimental Settings} \label{appendix:experimental-settings}
\subsection{Pre-training}
We implement our model in PyTorch, building on the Detectron2~\cite{wu2019detectron2} framework. We initialize the base models with the public weights of SAM2-Hiera-Large\footnote{https://github.com/facebookresearch/sam2} and BEIT-3-Large\footnote{https://huggingface.co/YxZhang/evf-sam2-multitask}. The model is pre‑trained for {25 epochs} on the {COCO‑2017} training split (104K images)~\citep{lin2014coco}. We use the \emph{panoptic} annotations, which provide pixel‑accurate masks and category labels for all 132 \emph{thing} and \emph{stuff} classes. Training is conducted on a single NVIDIA A100 (80 GB) GPU with a batch size of 8. Optimization employs AdamW (learning rate 1e-4). A step decay scheduler drops the learning rate by a factor of 0.1 at 89\% and 96\% of the total iterations. Compared with recent generalist models, our recipe is markedly more data‑efficient (see Table~\ref{tab:train-data-details}).

\begin{table*}[h]
\centering
\scriptsize           
\renewcommand{\arraystretch}{1.15}
\setlength{\tabcolsep}{4pt}

\caption{A detailed list of training data for generalist models and OpenWorldSAM.  
O365: Objects365.  
OID: OpenImages Detection.  
VG: Visual Genome.  
INB: ImageNetBoxes.  
RefC: RefCOCO/\,+/g.}
\label{tab:train-data-details}

\resizebox{\textwidth}{!}{\begin{tabular}{lp{4.0cm}p{1.5cm}p{1.3cm}p{4.2cm}}
\toprule
\multirow{2}{*}{\textbf{Method}} &
  \multicolumn{2}{c}{\textbf{Train Data (grouped by annotation types)}} &
  \multicolumn{2}{c}{\textbf{Image Consumption}}\\
\cmidrule(lr){2-3}\cmidrule(lr){4-5}
 &   \textbf{Instance-level} & \textbf{Image-level} &
   \textbf{Batch Size} &
   \textbf{\#Epoch\,$\times$\,\#Image \;/\; Batch Size\,$\times$\,\#Iter}\\

\midrule
X-Decoder \cite{zou2023xdecoder}  &
  COCO, RefC & Cap4M & 32, 1024 &
  200M (50 Ep $\times$ 4M Img) \\

OpenSeeD \cite{zhang2023openseed}  &
  COCO, O365 & -- & 32, 64 &
  48M (30 Ep $\times$ 1.8M Img) \\

APE (B) \cite{shen2024ape}  &
  COCO, LVIS, O365, OID, VG, RefC & -- & 16 &
  17.28M (16 Bs $\times$ 1.08M Iter) \\

\midrule
OpenWorldSAM & COCO & -- & 8 & 2.50M (25 Ep $\times$ 0.104M Img) \\
\bottomrule
\end{tabular}}
\end{table*}

\subsection{Zero-Shot Evaluation}
We evaluate semantic, instance, and panoptic segmentation in a \emph{zero‑shot} setting.  For instance segmentation and panoptic segmentation, we apply confidence-score filtering to remove masks with scores below 0.7, followed by non‑maximum suppression (NMS) with IoU threshold 0.5 to remove duplicate detections and retain distinct object instances. The confidence scores, originally termed ``estimated IoU scores'' in SAM~\cite{kirillov2023sam, ravi2024sam2}, are direct outputs from SAM2's mask decoder. These scores were optimized during SAM2's pre-training to select high-quality (\ie, confident) mask outputs.

\begin{table}[h]
\centering
\renewcommand{\arraystretch}{1.2}
\setlength{\tabcolsep}{6pt}
\caption{{Open-Vocabulary Segmentation Benchmark Statistics.}}
\resizebox{0.8\textwidth}{!}{\begin{tabular}{lcccccr}
\toprule
\multirow{2}{*}{\textbf{Evaluation Dataset}} & \multirow{2}{*}{\textbf{Scene type}} & \multicolumn{3}{c}{\textbf{Annotations}} & \multirow{2}{*}{\textbf{\# Images }} & \multirow{2}{*}{\textbf{\# Classes}}\\
 &  & Semantic & Instance & Panoptic &  \\
\midrule
ADE-150                & common   & \cmark & \cmark & \cmark & 2000 & 150 \\
ADE-847                & common   & \cmark & \xmark & \xmark & 2000 & 847 \\
Pascal Voc             & common   & \cmark & \xmark & \xmark         & 1449 & 20  \\
Pascal Context-59      & common   & \cmark & \xmark         & \xmark         & 5105 & 59  \\
Pascal Context-459     & common   & \cmark & \xmark         & \xmark         & 5105 & 459 \\
SUN RGB-D              & in-door  & \cmark & \xmark         & \xmark         & 5050 & 37  \\
ScanNet-20             & in-door  & \cmark & \xmark & \cmark         & 5436 & 20  \\
ScanNet-40             & in-door  & \cmark & \xmark & \xmark         & 5436 & 40  \\
\bottomrule
\end{tabular}}
\label{tab:openvocab_stats}
\end{table}

The open‑vocabulary benchmark comprises 5 datasets covering 8 different segmentation tasks; statistics are summarized in Table~\ref{tab:openvocab_stats}. We show a comprehensive evaluation protocol for open-vocabulary segmentation in various vocabulary sizes and image domains.

\subsection{Finetuning}
For referring‑expression segmentation we fine‑tune the pre‑trained checkpoint on RefCOCOg UMD training split for 10 epochs. Because images from RefCOCOg were seen during pre‑training (with category labels substituted for referring expressions ground truth), we adopt a conservative learning rate of 1e-5. We use a batch size of 8 during training.

\section{Qualitative Comparison on Two-Stage-Inference} \label{appendix:two-stage}
\begin{figure}[h]
    \centering
    \includegraphics[width=0.7\linewidth]{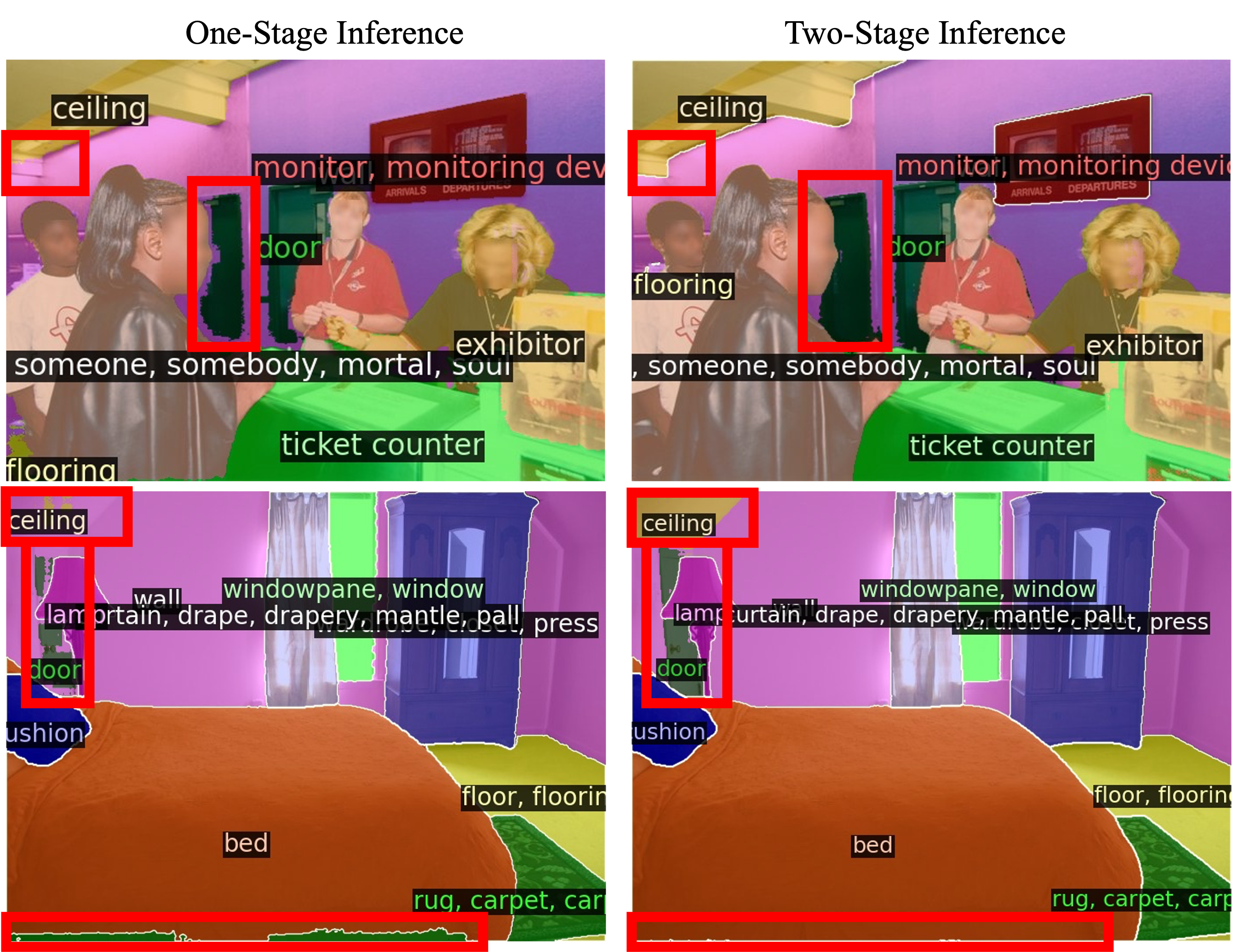}
    \caption{Qualitative results comparisons on using two-stage inference refinement on ADE20K-857.}
    \label{fig:two-stage-comparison}
\end{figure}

During inference, we perform an optional two-stage inference. First, the model predicts \emph{multi‑instance} masks. These masks are then fed back as visual prompts, and SAM2's mask decoder is run a second time to refine the contours. Figure~\ref{fig:two-stage-comparison} illustrates the visual improvement. However, quantitative gains are marginal across segmentation metrics (see Sec. 4.2 of the main paper), suggesting it mainly enhances visual quality rather than overall accuracy.  The reasons are twofold: (1) Two‑stage inference only refines mask contours; IoU‑style metrics saturate once coarse localization is accurate, so small contour tweaks seldom raise mIoU/PQ/AP; (2) Errors will be amplified on hard examples. On incorrectly localized masks from stage 1, refinement anchored to incorrect regions can further degrade metrics. Given that the two-stage inference serves as an optional, low-cost post-processing step, users can conveniently enable or disable it based on their preference.

\section{Additional Zero‑Shot Qualitative Results} \label{appendix:qualitative}

\begin{figure}[H]
\centering
\subfloat[ADE20K-150]{
  \includegraphics[width=\textwidth]{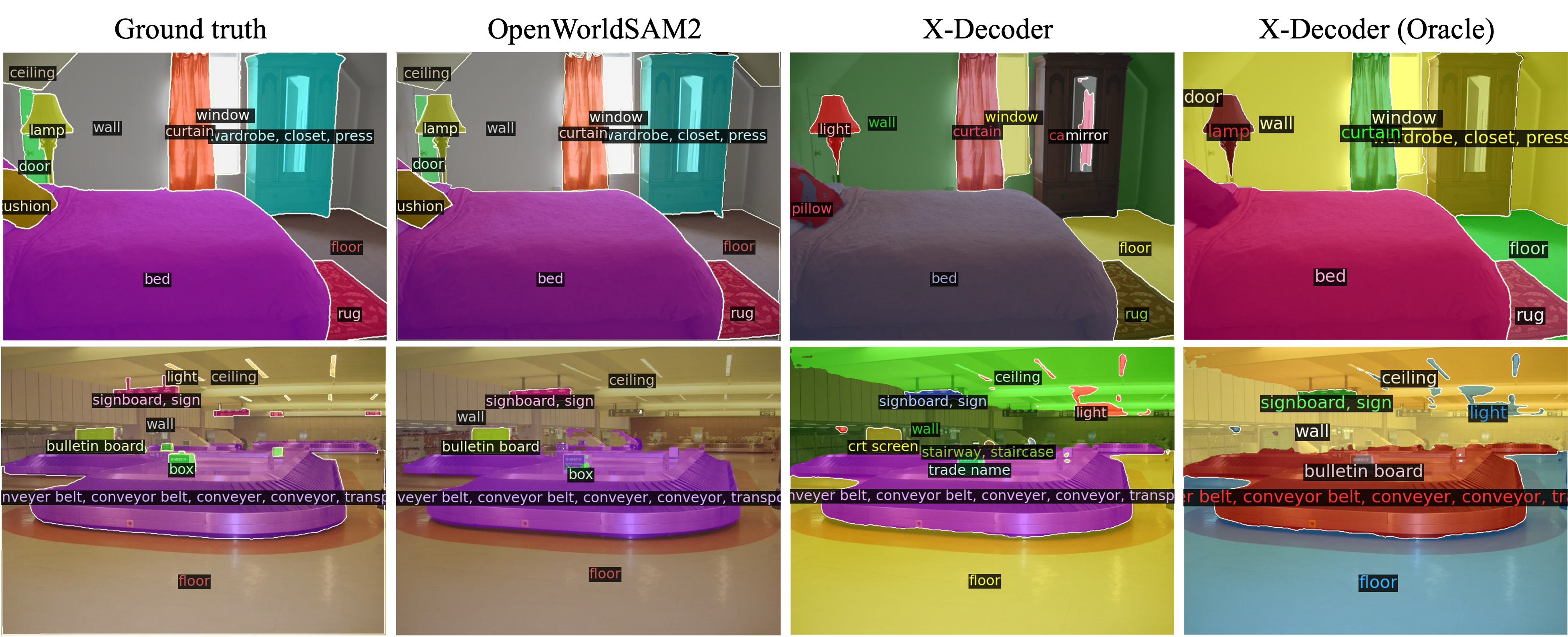}
}
\vspace{-10pt}
\subfloat[ADE20K-857]{
  \includegraphics[width=\textwidth]{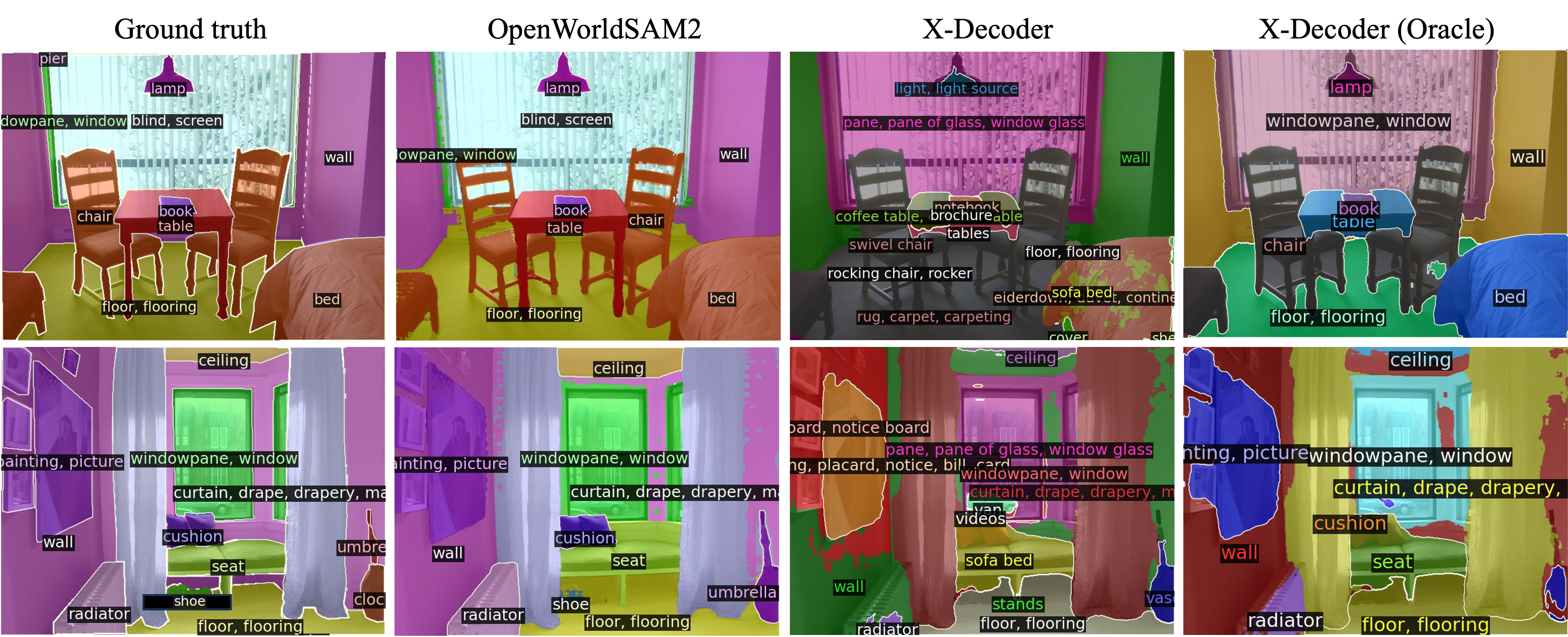}
}
\vspace{-10pt}
\subfloat[PASCAL Context-459]{
  \includegraphics[width=0.87\textwidth]{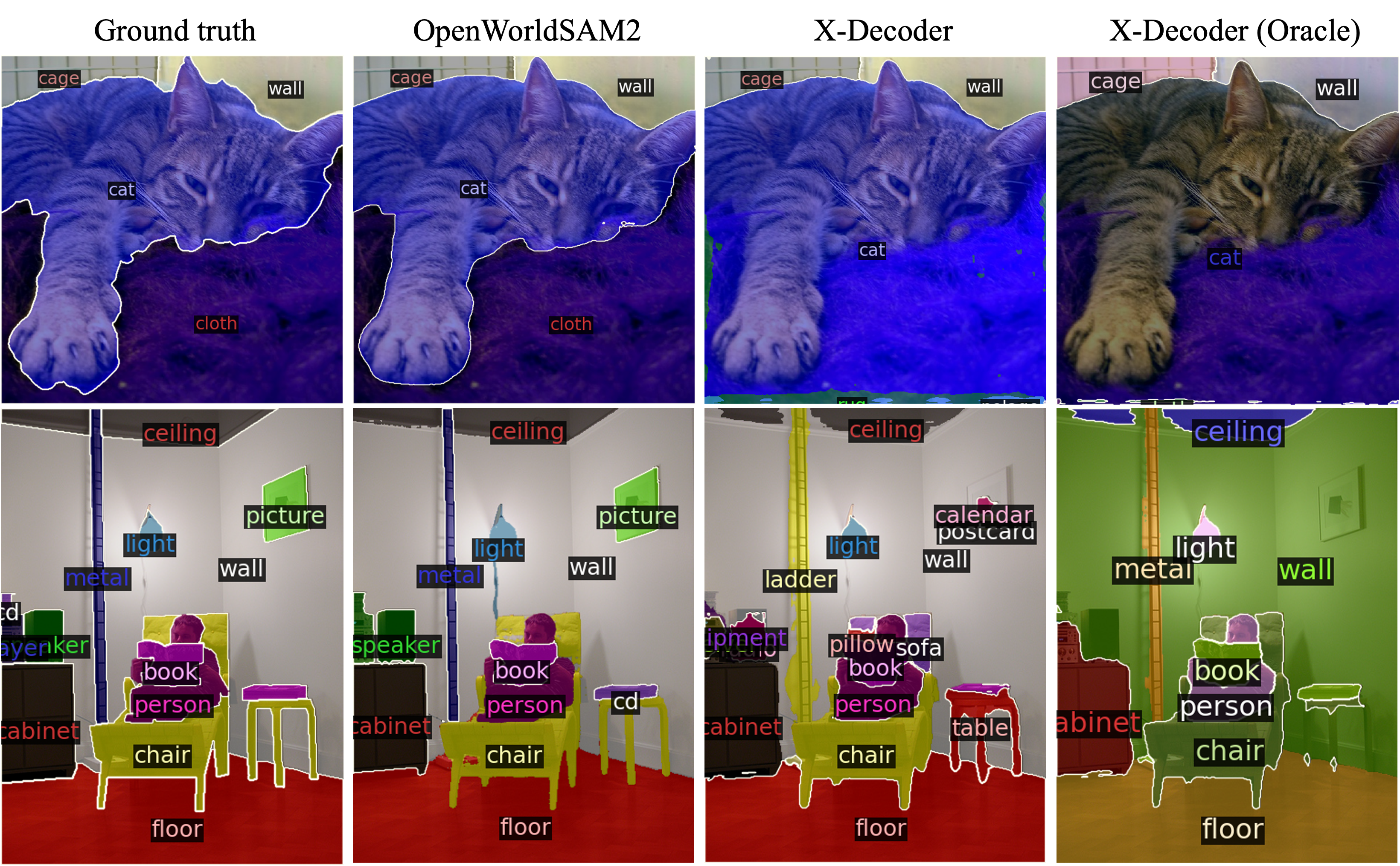}
}
\caption{{Qualitative comparisons between X-Decoder~\cite{zou2023xdecoder} and OpenWorldSAM on ADE20K-150, ADE20K-857, and PASCAL Context-459.}}
\label{fig:qualitative-more}
\end{figure}

Figure~\ref{fig:qualitative-more} showcases multiple challenging indoor scenes drawn from  ADE20K-150/857~\cite{zhou2017ade20k}, and PASCAL Context-459~\cite{mottaghi2014pascal-context}. In each sub-panel, we compare example outputs of OpenWorldSAM with comparisons to X-Decoder under both global-matching and oracle-prompts evaluation protocols.

\textit{Panel (a)} (ADE20K-150) top row depicts a cluttered bedroom. OpenWorldSAM cleanly delineates thin structures such as the ``closet’’ edge and the narrow ``lamp stem’’, and assigns a single coherent mask to the ``cushion''. X-Decoder fragments the closet and mis-classifies the cushion as a generic ``pillow’’ under global matching. Under oracle-prompts, X-decoder fails to predict ``cushion''. Similarly, the bottom row depicts an airport conveyor belt. X-Decoder mis-classifies the ``bulletin board'' as the ``crt screen'', the ``box'' as the ``trade name'' under global matching, and still mis-classifies the ``box'' under oracle-prompts.

\textit{Panel (b)} (ADE20K-857) top row shows a dining area. Under the global-matching protocol, X-Decoder hallucinates ``rug’’/``rocking chair'' labels and fragments the ``sofa bed'' pixels. The bottom row shows a cluttered living room where X-Decoder outputs fragmented low-quality masks and false predictions under both evaluation protocols. In comparison, our model preserves category fidelity—introducing no extra labels—and produces noticeably cleaner chair boundaries, illustrating the synergy between BEiT-3 language grounding and SAM2’s high-resolution masks.

\textit{Panel (c)} (PASCAL Context-459) top row shows that X-Decoder fails to predict the ``cloth'' object. The bottom row is an indoor scene crowded with small objects (``cd'', ``speaker'', ``chair''). OpenWorldSAM retrieves almost every queried category (except for ``cd'') and suppresses false positives such as ``{calendar}’’ and ``{ladder}’’ that appear in X-Decoder’s output, demonstrating stronger open-vocabulary grounding and sharper instance separation.

\section{Limitations - Outdoor Generalization } \label{appendix:limitation}


Despite strong results on indoor and everyday photographs, OpenWorldSAM under-performs on driving datasets such as Cityscapes \cite{cordts2015cityscapes} and BDD10K \cite{yu2018bdd100k} (Table \ref{tab:cityscapes}). Fine-tuning on Cityscapes narrows the gap, yet performance still trails methods explicitly exposed to multi-domain data. Understanding the source of this shortfall is essential for future extensions.

\begin{figure}[h]
    \centering
    \includegraphics[width=0.8\linewidth]{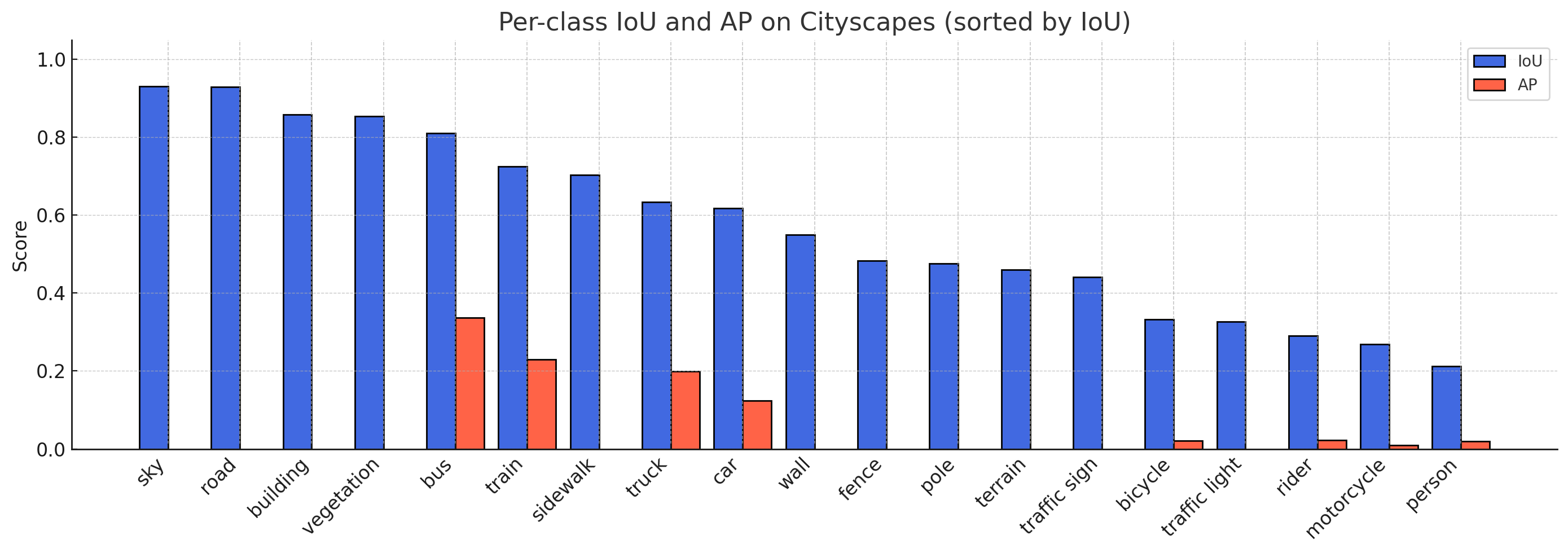}
    \caption{Per-class IoU And AP on Cityscapes (sorted by IoU). Performance collapses on thin or distant \emph{thing} classes (e.g.\ \emph{person}, \emph{traffic light}).}
    \label{fig:cityscapes}
\end{figure}

\textbf{Observed failure modes.} Figure \ref{fig:cityscapes} shows high IoU for broad \emph{stuff} regions (e.g.\ \emph{road}, \emph{sky}), but a sharp drop for small or elongated \emph{thing} instances. Correspondingly, AP remains low for \emph{motorcycle}, \emph{person}, \emph{bicycle}, etc.

\begin{table}[h]
    \centering
    \small
    \renewcommand{\arraystretch}{1.15}
    \setlength{\tabcolsep}{6pt}
    \caption{\textbf{Outdoor performance.} Open-vocabulary models are evaluated zero-shot on Cityscapes and BDD10K; the last row is fine-tuned on Cityscapes.}
    \begin{tabular}{llccccc}
    \toprule
       \multirow{2}{*}{\textbf{Model}}  & \multirow{2}{*}{\textbf{Evaluation}} & \multicolumn{3}{c}{\textbf{Cityscapes}} & \multicolumn{2}{c}{\textbf{BDD10K}} \\
         & & mIoU & AP & PQ & mIoU & PQ\\
    \midrule
         X-Decoder (L)~\cite{zou2023xdecoder}	& zero-shot & 52.0 & \textbf{24.9} & \textbf{38.1} & 	 \textbf{47.2} &  \textbf{17.8}\\
    \midrule
         OpenWorldSAM & zero-shot & 39.4 & 10.1 & 26.4 &	31.3 & 15.6 \\
         OpenWorldSAM & Finetune on Cityscapes &  \textbf{57.4} & 12.0 & 36.1 & 38.0 & 17.4\\
    \bottomrule
    \end{tabular}
    \label{tab:cityscapes}
\end{table}

\textbf{Hypotheses.}
\begin{enumerate}[label=(\roman*), leftmargin=1.3em, itemsep=0.2em]
    \item \textit{Domain shift.} COCO images are mostly handheld and indoor, whereas Cityscapes/BDD10K contain forward-looking dash-cam frames with motion blur, glare and night scenes. X-Decoder was co-trained on web-scale image-text pairs that include many outdoor photos, so its visual encoder has wider coverage. Large-scale multi-domain training is known to mitigate domain shift~\cite{bevandic2022multi}.
    \item \textit{Resolution bottleneck.} Cityscapes frames are $2048{\times}1024$. Rescaling to $1024{\times}1024$ (SAM default) reduces poles and traffic lights to nearly one pixel at the feature stride of $16\times$. X-Decoder keeps an FPN branch at $8\times$, preserving thin structures.
\end{enumerate}

\paragraph{Take-away.} COCO-only pre-training for OpenWorldSAM leaves a  blind spot for urban driving imagery—particularly for distant, thin or cluttered objects under challenging lighting. Bridging the gap likely requires (i) explicit exposure to outdoor domains and (ii)  higher-resolution feature branches. We leave large-scale outdoor pre-training and depth-aware augmentation for future work.

\section{Model Structure Details} \label{appendix:model-structure-details}
Table \ref{tab:model-configs} summarizes the architectural differences between OpenWorldSAM and competing models, detailing each method’s visual backbone, segmentation head, text encoder, and training-image resolution.

\begin{table}[h]
\centering
\small
\renewcommand{\arraystretch}{1.15}
\setlength{\tabcolsep}{6pt}
\caption{Architectural choices for recent open-vocabulary models: visual backbone, base segmentation model, text/multi-modal encoder, and training image size.}
\label{tab:model-configs}
\begin{tabular}{llcccc}
\toprule
\multirow{2}{*}{\textbf{Method}} & \multirow{2}{*}{\textbf{Visual Backbone}} & \multirow{2}{*}{\textbf{Base Model}} & \multirow{2}{*}{\textbf{Text Encoder}} & \multicolumn{2}{c}{\textbf{Image Size}} \\
\cmidrule(lr){5-6}
& & & & Short & Long \\
\midrule
MSeg (B)\,\cite{lambert2020mseg}     & HRNet-W48 (65 M)   & HRNet-Seg      &  --     & 1024 & 1024 \\
GroupViT (S)\,\cite{xu2022groupvit} & ViT-S/16 (22 M)    & GroupViT       &  Transformer & 224  & 224  \\
LSeg+ (B)\,\cite{li2022lseg}        & CLIP ViT-B/16 (86 M) & DenseCLIP     &  CLIP   & 512  & 512  \\
ZegFormer (B)\,\cite{ding2022zegformer} & CLIP ViT-B/16 (86 M) & ZegFormer  &  CLIP   & 640  & 640  \\
OpenSeg (B)\,\cite{ghiasi2022openseg}& ResNet-101 (45 M)  & OpenSeg        &  CLIP/ALIGN & 640 & 640 \\
OVSeg (B)\,\cite{liang2023ovseg}    & CLIP ViT-B/16 (86 M) & MaskFormer    &  CLIP   & 640  & 640  \\
MaskCLIP (L)\,\cite{dong2023maskclip} & CLIP ViT-L/14 (307 M) & MaskCLIP   &  CLIP   & 1024 & 1024 \\
X-Decoder~\cite{zou2023xdecoder}        & DaViT-L (196 M)    & X-Decoder   & CLIP   & 1024       & 1024 \\
OpenSeeD~\cite{zhang2023openseed}         & Swin-L (197 M)     & MaskDINO      & UniCL   & 1024            & 1024 \\
SEEM~\cite{zou2023seem}                 & DaViT-L (196 M)    & X-Decoder     & CLIP   & 800             & 1333 \\
APE (B)~\cite{shen2024ape}               & ViT-L (307 M)      & DETA          & CLIP    & 1024            & 1024 \\
PolyFormer (L)~\cite{liu2023polyformer}& Swin-L (197 M) & PolyFormer  & BERT           & 1024 & 1024 \\
UNINEXT (H)~\cite{yan2023universal}   & ViT-H (632 M)           & DINO        & BERT           & 320$\sim$800 & 1333 \\
PixelLM~\cite{ren2024pixellm}         & CLIP ViT-L/14 (307 M)       & PixelLM     & LLaMA2-13B     & 448 & 448 \\
LISA~\cite{lai2024lisa}               & SAM ViT-H (636 M)    & SAM       & Vicuna-7B      & 1024  & 1024  \\
GLaMM~\cite{rasheed2024glamm}         & SAM ViT-H (636 M)   & SAM       & Vicuna-7B      & 1024 & 1024 \\
u-LLaVA~\cite{xu2024ullava}           & SAM ViT-H (636 M)   & SAM     & Vicuna-7B      & 1024 & 1024 \\
EVF-SAM~\cite{zhang2024evfsam}        &  SAM ViT-H (636 M)        & SAM        & BEiT-3         & 1024 & 1024 \\
EVF-SAM2~\cite{zhang2024evfsam}        & SAM2 Hiera-L (224 M)       & SAM2        & BEiT-3         & 1024 & 1024 \\
\midrule
OpenWorldSAM & SAM2 Hiera-L (224 M) & SAM2 & BEiT-3 & 1024 & 1024\\
\bottomrule
\end{tabular}
\end{table}

\subsection{Possible Text Encoder Alternatives}
We argue that the key ingredients for open‑vocabulary segmentation are backbone‑agnostic: any strong interactive segmenter can supply high‑resolution mask decoding, while any pretrained vision‑language encoder can provide semantics.  What is missing is a lightweight adaptor that (i) aligns the two embedding spaces, (ii) scales to multiple object instances from a single text query, and (iii) preserves the  efficiency that makes interactive segmentation attractive in the first place. 

Our OpenWorldSAM is a general plug‑in architecture that satisfies these desiderata while keeping all heavy backbones frozen.  Although we instantiate the framework with SAM2 and BEiT‑3 in this paper, neither component is required by design; alternative interactive decoders or vision‑language encoders can be swapped in with only minor re‑training of the adapter.

Table~\ref{tab:vl-encoders} surveys representative VLM encoders that could replace BEiT-3 in OpenWorldSAM with {\small$\leq$}5\,M adaptor parameters.  All rows assume the heavy backbone is \textbf{frozen}; only the $256$-D projector and tie-breakers are fine-tuned.

\textbf{Adaptor fine-tuning recipe (all encoders).} Freeze all VLM weights and SAM2 decoder; initialize a $d_{\text{in}}\!\times\!256$ MLP projector and $K$ 256-D tie-breaker embeddings (default $K=20$, total ${\approx}5$M params).
For training, one could use unchanged Hungarian matching loss on COCO.

\textbf{Takeaway.} Early-fusion encoders (VLMo, OFA, Florence-2) require zero architectural change beyond projector resizing and are therefore the most promising immediate swaps.  Dual-encoders (CLIP family) need a shallow cross-attention adaptor to overcome missing image context.  Larger hybrids (BLIP-2, Kosmos-2, PaLI) open research directions (multi-query tie-breakers, OCR) at the cost of real-time guarantees.

\begin{sidewaystable}
\centering
\scriptsize
\renewcommand{\arraystretch}{1.15}
\setlength{\tabcolsep}{5pt}
\caption{Candidate vision–language encoders. ``TFM'' stands for Transformer. “Pooled dim” is the size of the single semantic vector exposed to the adaptor; “GFLOPs/Img” computed at $224^{2}$ resolution for the visual branch.}
\label{tab:vl-encoders}
\begin{tabular}{llccp{4.7cm}p{4.9cm}c}
\toprule
\textbf{Family / Exemplars} & \textbf{Arch.\ type} & \textbf{Pooled dim} & \textbf{Params} & \textbf{Pros for OpenWorldSAM} & \textbf{Adaptor–specific tweaks} & \textbf{GFLOPs/Img}\\
\midrule
\multicolumn{7}{c}{\em Early-fusion Transformers (drop-in closest to BEiT-3)}\\
\midrule
VLMo-B/L~\cite{bao2022vlmo}           & joint enc.\ TFM & 768  & 230/341M & same interface as BEiT-3; smaller model; multilingual & closest to BEiT-3 $\rightarrow$ just replace tokenizer + dimension in the projector; keep tie-breakers unchanged & 18.6/25.4 \\
OFA-B/L~\cite{wang2022ofa}  & joint enc.\ TFM & 768  & 184/312M & instruction-tuned; handy if we ever expose captioning & adjust tokenizer and change input dim in projector; reports slightly weaker alignment than BEiT-3 & 17.9/24.7\\
Florence-2-Base~\cite{xiao2024florence} & joint enc.\ TFM & 1024 & 230M & SOTA zero-shot retrieval; 10-lang support & none beyond changing tokenizer and input dim in projector  & 26.3\\
\midrule
\multicolumn{7}{c}{\em Dual-encoder Contrastive (text vector \emph{not} image-conditioned)}\\
\midrule
CLIP-ViT-L/14~\cite{radford2021clip}   & ViT+Text enc.\ & 768 & 304M & unlimited vocabulary; tiny latency;  many open checkpoints  & semantic vector is not image-conditioned $\rightarrow$ our ablation saw weaker performance. Mitigation: add a 2-layer cross-attn adapter that re-injects image tokens before the projector; expect AP drop if no cross-attn & 19.0\\
EVA-CLIP-E~\cite{sun2023eva}          & ViT-G/14 + Text & 1024 & 610M & stronger semantics than CLIP-L &  memory heavy; expect AP drop if no cross-attn & 37.2\\
SigLIP-2-S~\cite{tschannen2025siglip}      & ViT\,/\,Text & 512 & 86M & edge-friendly; multilingual &  expect AP drop if no cross-attn & 8.1\\
\midrule
\multicolumn{7}{c}{\em Encoder–Decoder w/ Contrastive Head (pooled vector from decoder)}\\
\midrule
CoCa-Base~\cite{yu2022coca}            & ViT enc.\ + TFM dec.\ & 768 & 365 M & better long-tail semantics & need to tap the unimodal decoder hidden state & 23.7\\
PaLI-3B~\cite{chen2022pali}            & ViT-E enc.\ + T5 dec.\ & 1024 & 3.0 B & 100-lang OCR; robust semantics & memory heavy; need to tap the unimodal decoder hidden state & 56.4\\
\midrule
\multicolumn{7}{c}{\em Query-former Hybrids (multiple vectors)}\\
\midrule
BLIP-2-OPT-2.7B~\cite{li2023blip}     & ViT + Q-Former + LLM & 32$\times$256 & 1.1B & native multi-query  & pool/average queries or extend SAM prompt len.\ & 31.5\\
Kosmos-2~\cite{peng2023kosmos}        & ViT enc.\ + LLM dec.\ & 768 & 1.6B & optional box tokens for UX studies & Requires a one-step decode per prompt (latency) and an additional MLP to strip location bias & 34.8\\
\bottomrule
\end{tabular}
\end{sidewaystable}

\section{Additional Ablation Studies}
We provide additional ablation studies on the number of tie-breaker tokens and the number of cross-attention layers.

\subsection{Effect of varying tie-breaker tokens}
We set the hyperparameter $K=20$, meaning for each prompt (e.g., a category name), our model can identify up to 20 distinct objects. For crowded scenes containing more than 20 objects per category, increasing $K$ is straightforward and advisable. In practice, COCO images typically contain a moderate number of distinct categories and instances (the original COCO paper reports ``on average, our dataset contains 3.5 categories and 7.7 instances per image.''~\cite{lin2014coco}). The chosen value should match or exceed the maximum expected number of objects per category. For reference, DETR~\cite{carion2020detr} used 100 total queries, aligning roughly with the maximum number of objects per image. Our choice ($K=20$) results, on average, in approximately 70 queries per image (20 queries $\times$ 3.5 categories), providing ample coverage for typical scenes.

Further, \cite{zhang2022expected} observed that increasing queries initially improved Average Precision (AP), but then plateaued or even slightly declined when queries became excessive, indicating redundancy in higher query counts. However, recall does improve with more queries, since more detection slots increase the chance to find each object.

We conducted additional ablation experiments in Table \ref{tab:tie-breaker-num} by varying $K$, pretrained on COCO and evaluated on ADE20K instance segmentation.
\begin{table}[H]
\centering
\scriptsize
\caption{Ablation on the number of tie-breakers $K$.}
\label{tab:tie-breaker-num}
\setlength{\tabcolsep}{6pt}
\renewcommand{\arraystretch}{1.15}
\resizebox{0.6\textwidth}{!}{%
\begin{tabular}{lccc}
\toprule
\textbf{Metric} & \textbf{$K=10$} & \textbf{$K=20$} & \textbf{$K=30$} \\
\midrule
Average Precision (AP)	&14.2&	\textbf{16.9}	&16.5\\
Average Recall@100 (AR)	&21.6	&28.8&	\textbf{29.4}\\
\bottomrule
\end{tabular}%
}
\end{table}

Observations. (1) Increasing $K$ from 10 to 20 improves recall and AP; beyond 20 gains saturate, mirroring the behavior reported for DETR‑style object queries; (2) Average Recall with max 100 detections per image (AR@100) improve when increasing $K$
 from $10 \rightarrow 20  \rightarrow 30$; (3) $K=20$
 is optimal for balancing precision and recall in standard datasets.

 \subsection{Effect of varying number of cross-attention layers}

 In Table \ref{tab:crossattn-ablation}, we observe consistently higher accuracy with 3-layer cross-attention across datasets, confirming the importance of multi-layer cross-attention. However, a single-layer variant significantly narrows the gap with fewer parameters (2.4M vs. 4.5M), suggesting a practical compromise between parameter count and accuracy. 

 \begin{table}[h]
\centering
\scriptsize
\caption{Ablation on cross-attention depth across datasets. Metrics are PQ/AP/mIoU for ADE-150 and mIoU for the others.}
\label{tab:crossattn-ablation}
\setlength{\tabcolsep}{4.5pt}
\renewcommand{\arraystretch}{1.15}
\resizebox{\textwidth}{!}{%
\begin{tabular}{lcccccccc}
\toprule
\textbf{Variant} & \textbf{Params} & \textbf{ADE-150 (PQ/AP/mIoU)} & \textbf{ADE-857 (mIoU)} & \textbf{PC-59 } & \textbf{PC-459 } & \textbf{VOC-20 } & \textbf{SUN-37 } & \textbf{SCAN-40 } \\
\midrule
no cross-attn           & 1.7 (M) & 35.1\,/\,17.1\,/\,56.8 & 32.2 & 70.4 & 44.2 & 97.3 & 63.6 & 53.8 \\
1-layer cross-attn  & 2.4  (M)& 35.1\,/\,16.8\,/\,59.0 & 32.8 & 72.6 & 46.3 & 97.5 & 66.4 & 54.0 \\
3-layer cross-attn                & 4.5 (M) & 35.2\,/\,16.9\,/\,60.4 & 33.1 & 73.7 & 47.5 & 98.0 & 67.7 & 55.6 \\
\bottomrule
\end{tabular}%
}
\end{table}

\section{Inference Speed Analysis}
We conducted detailed profiling to quantify the impact of adding the VLM and our adapter modules to SAM. In Table \ref{tab:profiling-single} and \ref{tab:profiling-six}, we present inference timing breakdowns for processing a single $1024\times1024$ image on an NVIDIA A5000 GPU, averaged over five independent runs.

\begin{table}[h]
\centering
\scriptsize
\caption{Inference timing breakdown for a single text prompt (20 queries).}
\label{tab:profiling-single}
\setlength{\tabcolsep}{6pt}
\renewcommand{\arraystretch}{1.15}
\resizebox{0.7\textwidth}{!}{%
\begin{tabular}{lrrl}
\toprule
\textbf{Module} & \textbf{Time (ms)} & \textbf{Percentage} & \textbf{Category} \\
\midrule
\texttt{sam\_backbone\_feature\_prep} & 329.83 & 71.6\% & SAM \\
\texttt{prompt\_tokenization} & 0.43 & 0.1\% & Non\text{-}SAM \\
\texttt{beit3\_forward} & 70.84 & 15.4\% & Non\text{-}SAM \\
\texttt{mlp\_projection\_layer} & 6.68 & 1.4\% & Non\text{-}SAM \\
\texttt{prepare\_batched\_tie\_breaker\_tokens} & 0.13 & 0.0\% & Non\text{-}SAM \\
\texttt{cross\_attention} & 8.45 & 1.8\% & Non\text{-}SAM \\
\texttt{sam\_prompt\_encoder} & 0.11 & 0.0\% & SAM \\
\texttt{sam\_mask\_decoder} & 43.41 & 9.4\% & SAM \\
\texttt{postprocessing} & 0.68 & 0.1\% & Non\text{-}SAM \\
\midrule
\textbf{TOTAL TIME} & \textbf{460.69} & \textbf{100.0\%} & \textemdash \\
\bottomrule
\end{tabular}%
}
\end{table}

\begin{table}[h]
\centering
\scriptsize
\caption{Inference timing breakdown for six text prompts (120 queries).}
\label{tab:profiling-six}
\setlength{\tabcolsep}{6pt}
\renewcommand{\arraystretch}{1.15}
\resizebox{0.7\textwidth}{!}{%
\begin{tabular}{lrrl}
\toprule
\textbf{Module} & \textbf{Time (ms)} & \textbf{Percentage} & \textbf{Category} \\
\midrule
\texttt{sam\_backbone\_feature\_prep} & 334.42 & 48.6\% & SAM \\
\texttt{prompt\_tokenization} & 1.02 & 0.1\% & Non\text{-}SAM \\
\texttt{beit3\_forward} & 123.73 & 18.0\% & Non\text{-}SAM \\
\texttt{mlp\_projection\_layer} & 4.48 & 0.6\% & Non\text{-}SAM \\
\texttt{prepare\_batched\_tie\_breaker\_tokens} & 0.20 & 0.0\% & Non\text{-}SAM \\
\texttt{cross\_attention\_layers} & 18.17 & 2.6\% & Non\text{-}SAM \\
\texttt{sam\_prompt\_encoder} & 0.12 & 0.0\% & SAM \\
\texttt{sam\_mask\_decoder} & 205.18 & 29.8\% & SAM \\
\texttt{postprocessing} & 1.06 & 0.2\% & Non\text{-}SAM \\
\midrule
\textbf{TOTAL TIME} & \textbf{688.50} & \textbf{100.0\%} & \textemdash \\
\bottomrule
\end{tabular}%
}
\end{table}

\noindent\textbf{Summary (single prompt).} SAM modules total time: 373.35\,ms (81.0\%),
Non\text{-}SAM modules total time: {87.21\,ms} (18.9\%), Non\text{-}SAM overhead: {87.21\,ms}.

\noindent\textbf{Summary (six prompts).} SAM modules total time: {539.72\,ms} (78.4\%), Non\text{-}SAM modules total time: {148.65\,ms} (21.6\%), Non\text{-}SAM overhead: {148.65\,ms}.

\paragraph{Takeaway.}
The profiling results show that adding the VLM and adapter modules results in only a moderate increase in inference time (approximately 19--22\% overhead). Most computational cost remains within SAM's backbone and mask decoder.

\paragraph{Mask Decoder scaling.}
\texttt{sam\_mask\_decoder} cost grows almost linearly with $(K\times P)$.
\begin{itemize}
\item Going from $1\rightarrow 20$ queries (same prompt) adds $\sim$41\,ms.
\item Going from 1 prompt $\rightarrow$ 6 prompts (120 queries) adds a further $\sim$162\,ms.
\end{itemize}
Note that one text prompt mimics user clicks 20 times on an image. If automatic mask generation is desired without user intervention, SAM's built-in auto-mask generator uses a dense $32\times32$ grid of point prompts, incurring significantly higher costs compared to our text-based prompting approach.

\paragraph{Overall overhead.}
Relative to one vanilla SAM2 call, our pipeline is approximately 39\% slower for a single prompt (332\,$\rightarrow$\,461\,ms). However, it becomes approximately $3\times$ more efficient when handling three or more prompts, as the backbone and VLM overhead are amortized. Thus, our enhancements introduce manageable overhead, maintaining practical usability in real-world applications.

\end{document}